\documentclass[%
 aip,
 amsmath,amssymb,
 reprint,%
]{revtex4-1}

\usepackage{graphicx}% Include figure files
\usepackage{dcolumn}% Align table columns on decimal point
\usepackage{bm}% bold math
\usepackage[utf8]{inputenc}
\usepackage[T1]{fontenc}
\usepackage{mathptmx}
\usepackage{etoolbox}

\usepackage{amsmath}
\usepackage{accents}

\usepackage{hyperref}
\usepackage{xurl}

\newlength{\dhatheight}

\def\x{\mathbf{x}}

\def\y{\mathbf{y}}

\def\Bmu{\bm{\mu}}
\def\Bsigma{\bm{\sigma}}

\makeatletter
\def\@email#1#2{%
 \endgroup
 \patchcmd{\titleblock@produce}
  {\frontmatter@RRAPformat}
  {\frontmatter@RRAPformat{\produce@RRAP{*#1\href{mailto:#2}{#2}}}\frontmatter@RRAPformat}
  {}{}
}%
\makeatother
\begin{document}

\preprint{AIP/123-QED}

%\title[Bidirectional Autoregressive Latent Diffusion for Magnetohydrodynamics]{Bidirectional Autoregressive Latent Diffusion for Magnetohydrodynamics}
\title[Bidirectional Autoregressive Latent Diffusion for Forward and Inverse Magnetohydrodynamics]{Bidirectional Autoregressive Latent Diffusion \\for Forward and Inverse Magnetohydrodynamics}
% Force line breaks with \\
\author{Alexander Scheinker}
 \affiliation{Los Alamos National Laboratory, Los Alamos, NM, 87545.}%Lines break automatically or can be forced with \\
 \email{ascheink@lanl.gov}

\date{\today}% It is always \today, today,
             %  but any date may be explicitly specified

\begin{abstract}
This work presents a new bidirectional autoregressive latent diffusion approach for predicting the evolution of multiple fields (mass density, pressure, velocity, and magnetic field components) for magnetohydrodynamics. We show that this bidirectional flow can be used as a self-supervised consistency metric for uncertainty and error estimation, which enables the model to estimate test-time uncertainty and error without access to ground truth, by comparing how closely flowing forwards and backwards in time returns to the same predicted fields. We also demonstrate this method’s potential to serve as a non-invasive plasma diagnostic, and show how adaptive feedback can be used to make the model more robust based on sparse diagnostics or limited views/measurements.
\end{abstract}

\maketitle

%%%%%%%%%%%%%%%%%%%%%%%%%%%%%%%%%%%%%%%%%%%%%%%%%%%%%%%%%%%%%%%%%%
%%%%%%%%%%%%%%%%%%%%%%%%%%%%%%%%%%%%%%%%%%%%%%%%%%%%%%%%%%%%%%%%%%
%%%%%%%%%%%%%%%%%%%%%%%%%%%%%%%%%%%%%%%%%%%%%%%%%%%%%%%%%%%%%%%%%%
%%%%%%%%%%%%%%%%%%%%%%%%%%%%%%%%%%%%%%%%%%%%%%%%%%%%%%%%%%%%%%%%%%
\section{Introduction}\label{sec:intro}
%%%%%%%%%%%%%%%%%%%%%%%%%%%%%%%%%%%%%%%%%%%%%%%%%%%%%%%%%%%%%%%%%%
%%%%%%%%%%%%%%%%%%%%%%%%%%%%%%%%%%%%%%%%%%%%%%%%%%%%%%%%%%%%%%%%%%
%%%%%%%%%%%%%%%%%%%%%%%%%%%%%%%%%%%%%%%%%%%%%%%%%%%%%%%%%%%%%%%%%%
%%%%%%%%%%%%%%%%%%%%%%%%%%%%%%%%%%%%%%%%%%%%%%%%%%%%%%%%%%%%%%%%%%

Magnetohydrodynamics (MHD) describes plasma dynamics modeled as an electrically conducting fluid under the influence of its own electromagnetic fields \cite{alfven1942existence}. MHD describes systems that include plasma turbulence \cite{matthaeus2021turbulence,cho2002compressible}, quantum
plasmas \cite{haas2005magnetohydrodynamic}, Tokamaks \cite{korving2023development,orain2013non,jardin2022ideal,izzo2008magnetohydrodynamic,heidbrink2008basic}, plasma detachment in magnetic nozzles \cite{arefiev2005magnetohydrodynamic}, liquid metals \cite{smolentsev2021physical,gill2007development}, and ion beams in particle accelerators \cite{baranov1991mhd}.

In this work we study the use of bidirectional autoregressive latent diffusion models that can flow forward or backback in time to predict the evolution of multiple fields (mass density, pressure, velocity, and magnetic field components) for magnetohydrodynamics. We start by creating and training with arbitrary mixture-of-Gaussian synthetic MHD densities and then test those models on the Orszag–Tang vortex, which is a widely used test case for 2D MHD simulations \cite{orszag1979small}.

We show that this bidirectional flow can be used as a self-supervised consistency check for our generative model for test-time estimation of prediction errors by comparing whether cycling forwards and backwards in time returns to the same predicted fields. We also demonstrate this method’s potential to serve as a non-invasive plasma diagnostic, and show how adaptive feedback can be used to make the model more robust based on sparse diagnostics/measurements.

Such generative models have the potential to provide accurate surrogates of MHD simulations which run much faster than traditional approaches, which could be useful for complex plasma processes that would benefit from real-time data processing \cite{choi2020data} and real-time data analysis \cite{kube2022near}. Although not as accurate as physics models, these methods have two main benefits in being very fast for real-time applications such as finding approximate solutions, and the ability to roll backward in time is useful for solving inverse problems for operational systems with limited direct observations by providing a good initial guess of the plasma's state. Overall, such an approach can potentially contribute to provide virtual plasma diagnostics which can be used for real-time feedback control for complex plasma physics systems. This would extend to plasmas work that is already underway for using adaptively tuned advanced physics-constrained latent diffusion-based generative models as real-time diagnostics for time-varying beams in charged particle accelerators based on limited views/measurements \cite{scheinker2026phaseflow4d}.

%%%%%%%%%%%%%%%%%%%%%%%%%%%%%%%%%%%%%%%%%%%%%%%%%%%%%%%%%%%%%%%%%%
%%%%%%%%%%%%%%%%%%%%%%%%%%%%%%%%%%%%%%%%%%%%%%%%%%%%%%%%%%%%%%%%%%
\subsection{\label{sec:level2}Summary of Main Contributions}
%%%%%%%%%%%%%%%%%%%%%%%%%%%%%%%%%%%%%%%%%%%%%%%%%%%%%%%%%%%%%%%%%%
%%%%%%%%%%%%%%%%%%%%%%%%%%%%%%%%%%%%%%%%%%%%%%%%%%%%%%%%%%%%%%%%%%

In this work, we present a new autoregressive latent diffusion model for predicting the evolution of multiple fields (mass density, pressure, velocity, and magnetic field components) for magnetohydrodynamics. Our main contributions are summarized as:
\begin{itemize}
    \item Ability to roll out predictions autoregressively both forward and backward in time, thus serving as both a surrogate model and an inverse problem solver that can recover initial and intermediate plasma states from final measured output conditions.
    \item Self-supervised consistency metric that allows the model to estimate test-time uncertainty and error without access to ground truth.
    \item Demonstration of how adaptive feedback coupled with this approach can be used to make the model more robust as a virtual plasma diagnostic based on sparse physical diagnostics or limited views/measurements.
\end{itemize}

\begin{figure*}[ht]
\centering
\includegraphics[width=1.0\linewidth]{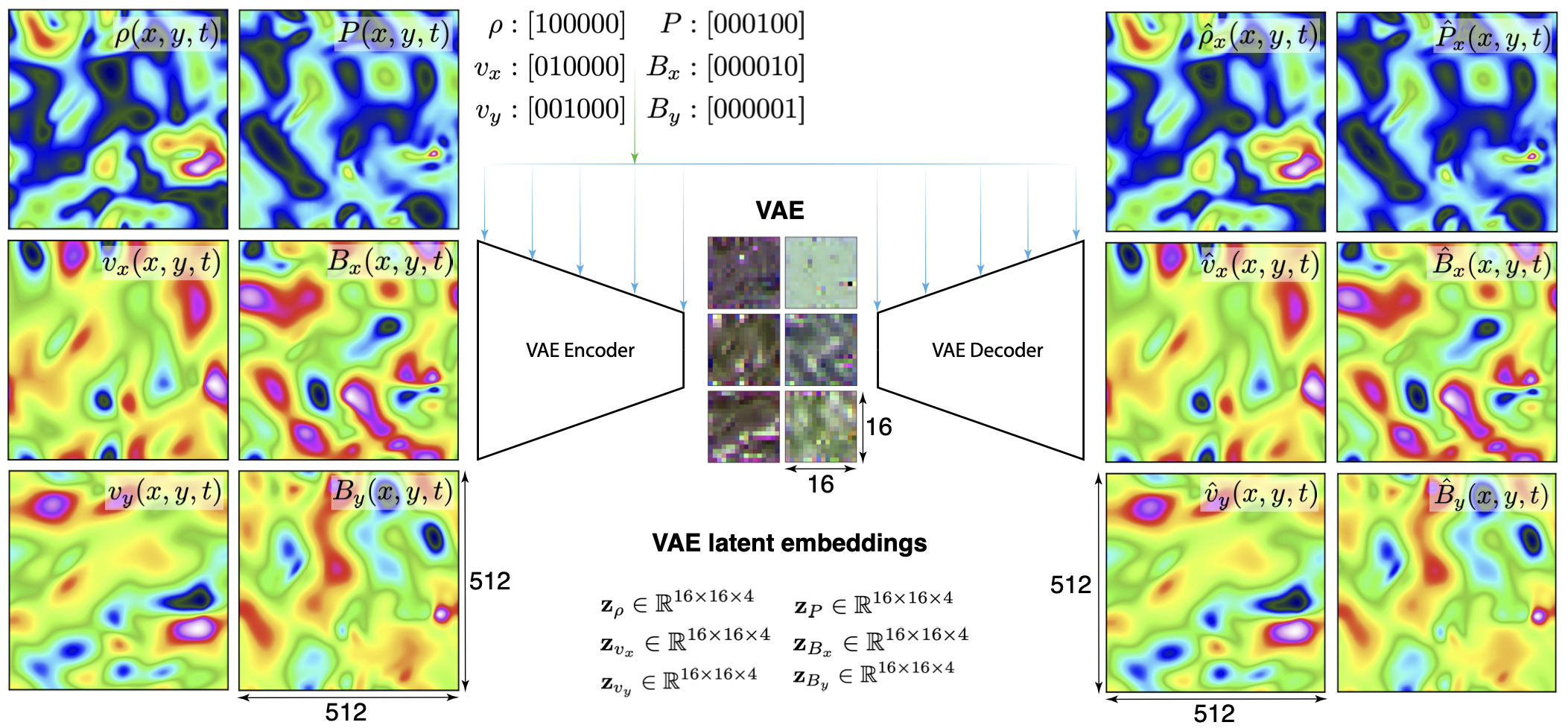}
\caption{Conditional diffusion overview. Each MHD field is encoded into its own lower-dimensional latent representation.}
\label{fig:setup}
\end{figure*}

%%%%%%%%%%%%%%%%%%%%%%%%%%%%%%%%%%%%%%%%%%%%%%%%%%%%%%%%%%%%%%%%%%
%%%%%%%%%%%%%%%%%%%%%%%%%%%%%%%%%%%%%%%%%%%%%%%%%%%%%%%%%%%%%%%%%%
%%%%%%%%%%%%%%%%%%%%%%%%%%%%%%%%%%%%%%%%%%%%%%%%%%%%%%%%%%%%%%%%%%
%%%%%%%%%%%%%%%%%%%%%%%%%%%%%%%%%%%%%%%%%%%%%%%%%%%%%%%%%%%%%%%%%%
\section{2D MHD Model}
%%%%%%%%%%%%%%%%%%%%%%%%%%%%%%%%%%%%%%%%%%%%%%%%%%%%%%%%%%%%%%%%%%
%%%%%%%%%%%%%%%%%%%%%%%%%%%%%%%%%%%%%%%%%%%%%%%%%%%%%%%%%%%%%%%%%%
%%%%%%%%%%%%%%%%%%%%%%%%%%%%%%%%%%%%%%%%%%%%%%%%%%%%%%%%%%%%%%%%%%
%%%%%%%%%%%%%%%%%%%%%%%%%%%%%%%%%%%%%%%%%%%%%%%%%%%%%%%%%%%%%%%%%%
For modeling 2D MHD, we utilize the tool and approach of P. Mocz \cite{mocz2014constrained, mocz2016moving, mocz2023ct_code, mocz2023ct_article}, which considers the conservation law dynamics
\begin{equation}
    \frac{\partial \mathbf{U}}{\partial t} + \nabla \cdot \mathbf{F}(\mathbf{U}) = 0,
\end{equation}
\begin{equation}
\mathbf{U} = \begin{pmatrix}
\rho \\
\rho\mathbf{v} \\
\rho e \\
\mathbf{B}
\end{pmatrix}, 
\quad 
\mathbf{F}(\mathbf{U}) = \begin{pmatrix}
\rho \mathbf{v} \\
\rho\mathbf{v}\mathbf{v}^T + p - \mathbf{B}\mathbf{B}^T\\
\rho e \mathbf{v} + p\mathbf{v} - \mathbf{B}(\mathbf{v}\cdot \mathbf{B}) \\
\mathbf{B} \mathbf{v}^T - \mathbf{v}\mathbf{B}^T,
\end{pmatrix}
\end{equation}
where $p=p_\mathrm{gas} + \frac{1}{2}\mathbf{B}^2$ is the total gass pressure, $e=u+\frac{1}{2}\mathbf{v}^2+\frac{1}{2\rho}\mathbf{B}^2$ is the total energy per unit mass, and $u$ is the thermal energy per unit mass.

For this work we generate synthetic 2D fields with randomized initial density, pressure, and velocity profiles and the magnetic fields are generated by first creating a random vector potential
which is a mixture of 100 Gaussians, from which the initial magnetic field components are then calculated. All of the simulations are run at a $512\times 512$-pixel resolution, with each simulation saving 100 time-steps over a 1-second run. We generate 500 such random MHD trajectories for training and 50 additional trajectories as test data.

%%%%%%%%%%%%%%%%%%%%%%%%%%%%%%%%%%%%%%%%%%%%%%%%%%%%%%%%%%%%%%%%%%
%%%%%%%%%%%%%%%%%%%%%%%%%%%%%%%%%%%%%%%%%%%%%%%%%%%%%%%%%%%%%%%%%%
%%%%%%%%%%%%%%%%%%%%%%%%%%%%%%%%%%%%%%%%%%%%%%%%%%%%%%%%%%%%%%%%%%
%%%%%%%%%%%%%%%%%%%%%%%%%%%%%%%%%%%%%%%%%%%%%%%%%%%%%%%%%%%%%%%%%%
\section{Latent Diffusion}
%%%%%%%%%%%%%%%%%%%%%%%%%%%%%%%%%%%%%%%%%%%%%%%%%%%%%%%%%%%%%%%%%%
%%%%%%%%%%%%%%%%%%%%%%%%%%%%%%%%%%%%%%%%%%%%%%%%%%%%%%%%%%%%%%%%%%
%%%%%%%%%%%%%%%%%%%%%%%%%%%%%%%%%%%%%%%%%%%%%%%%%%%%%%%%%%%%%%%%%%
%%%%%%%%%%%%%%%%%%%%%%%%%%%%%%%%%%%%%%%%%%%%%%%%%%%%%%%%%%%%%%%%%%

Although generative diffusion directly in image-space gives the highest accuracy results, working with such large $(512\times 512)$ images can make the iterative diffusion flow-based generative process very slow, both for training and for inference, because to generate data, the models learn to flow from an initial analytically known noise distribution $\mathcal{N}(\mathbf{0},I_{d\times d\times c})$ to their estimate $\hat{\rho}_\mathrm{data}$ of the probability distribution describing the provided data samples $\rho_\mathrm{data}$. Here $d\times d$ represents image size and $c$ is the number of image channels, which, for example, for RGB images is typically 3, or 4 if they include opacity. This flow takes the form of a finite-difference approximation of a stochastic diffusion differential equation, which requires tens to hundreds of steps for the generative process. For very large images, using a very large denoising diffusion model, this becomes very computationally expensive.

The latent diffusion approach is to first compress the data to a much smaller latent representation by using the encoder of a variational autoencoder (VAE), $\mathcal{E}_\theta$, which maps images $X\in\mathbb{R}^{d\times d \times c}$ down to latent representations $\mathcal{E}_\theta(X)=\mathbf{z}\in\mathbb{R}^{L_d \times L_d \times L_c}$ with $L_d \times L_d \times L_c \ll d\times d\times c$. The overall setup is shown in Figure \ref{fig:setup}.

%%%%%%%%%%%%%%%%%%%%%%%%%%%%%%%%%%%%%%%%%%%%%%%%%%%%%%%%%%%%%%%%%%
%%%%%%%%%%%%%%%%%%%%%%%%%%%%%%%%%%%%%%%%%%%%%%%%%%%%%%%%%%%%%%%%%%
\subsection{Variational Autoencoder}
%%%%%%%%%%%%%%%%%%%%%%%%%%%%%%%%%%%%%%%%%%%%%%%%%%%%%%%%%%%%%%%%%%
%%%%%%%%%%%%%%%%%%%%%%%%%%%%%%%%%%%%%%%%%%%%%%%%%%%%%%%%%%%%%%%%%%
In our approach, for any single field, $d\times d = 512\times 512 = 262144$, while $L_d \times L_d \times L_c = 16 \times 16 \times 4 = 1024$, resulting in a $256\times$ factor of data compression. The job of the VAE decoder, $\mathcal{D}_\theta$, is to then reconstruct the input as accurately as possible based only on the low-dimensional latent representation, resulting in an overall data flow of
\begin{equation}
    X \quad \underbrace{\xrightarrow{\hspace{1.0cm}}}_{\mathcal{E}_\theta}\quad \mathbf{z} \quad \underbrace{\xrightarrow{\hspace{1.0cm}}}_{\mathcal{D}_\theta} \quad \hat{X}.
\end{equation}
Decreasing latent space dimension increases compression amount, making the VAE's job more difficult and decreases the accuracy of reconstructions $\hat{X}$. Optimal latent dimension is unique to each data set, it depends on the image resolution, the intrinsic dimensionality of the manifold on which the data lives, and on the number of weights $\theta$ and the architecture of the VAE. In practice this needs to be tested for each new dataset. In this approach we found that going down to $\mathbf{z}\in\mathbb{R}^{8\times 8 \times 4}$ made it too difficult for our VAE to recover fine details of the fields, while keeping a larger $\mathbf{z}\in\mathbb{R}^{32\times 32 \times 4}$ only showed minor improvement relative to $\mathbf{z}\in\mathbb{R}^{8\times 8 \times 4}$.

The VAE's architecture is actually a little bit different from that of a standard autoencoder, rather than simply generating $\mathbf{z}=\mathcal{E}_\theta(X)$ directly, the encoder's output is actually split into two parts which are then used to define a normal distribution from which the latent representation is sampled before being passed to the decoder
\begin{eqnarray}
    && X \ \underbrace{\xrightarrow{\hspace{1.0cm}}}_{\mathcal{E}_\theta} \ [\Sigma_\theta, \mu_\theta]=\mathcal{E}_\theta(X),\\
    && \mathbf{z} = \mu_\theta + \Sigma_\theta \epsilon, \quad \epsilon \sim \mathcal{N}(\mathbf{0},I_{L_d \times L_d \times L_c}), \\
    && \mathbf{z} \ \underbrace{\xrightarrow{\hspace{1.0cm}}}_{\mathcal{D}_\theta} \ \hat{X} = \mathcal{D}_\theta(\mathbf{z}).
\end{eqnarray}
Note that compared to a standard autoencoder (AE), whose generative process takes the form
\begin{eqnarray}
    && X \ \underbrace{\xrightarrow{\hspace{1.0cm}}}_{\mathcal{E}_\theta} \ \mu_\theta=\mathcal{E}_\theta(X),\\
    && \mathbf{z} = \mu_\theta, \\
    && \mathbf{z} \ \underbrace{\xrightarrow{\hspace{1.0cm}}}_{\mathcal{D}_\theta} \ \hat{X} = \mathcal{D}_\theta(\mathbf{z}) = \mathcal{D}_\theta(\mathcal{E}_\theta(X)),
\end{eqnarray}
the VAE can be thought of as an AE in which even for a fixed input $X$, random noise is added to slightly perturb the latent representations $\mathbf{z}=\mathcal{E}_\theta(X)$, which makes the decoder's job more difficult as it must try and reconstruct the same $X$ from slightly perturbed latent representations. 

Another major difference is that for a VAE, the training loss is not only reconstruction error, but there is an additional Kullback-Leibler (KL)-divergence term which tries to force the latent space to look like $p(\mathbf{z})=\mathcal{N}(\mathbf{0},I_{L_d \times L_d \times L_c})$. VAE training loss is defined as
\begin{equation}
    \mathcal{L} = -\mathbb{E}_{q_\theta(\mathbf{z}|X)}\left[\log p_\theta(X|\mathbf{z})\right] + \beta\, D_{\mathrm{KL}}\!\left(q_\theta(\mathbf{z}|X)\,\|\,p(\mathbf{z})\right),
\end{equation}
where $q_\theta(\mathbf{z}|X)$ is the probability distribution of the encoder and $p(\mathbf{z})$ is the prior that we want to match. The KL divergence term $D_{KL}$ is defined in general as
\begin{equation}
    D_{\mathrm{KL}}\!\left(q_\theta(\mathbf{z}|X)\,\|\,p(\mathbf{z})\right) = \int q_\theta(\mathbf{z}|X)\,\log\frac{q_\theta(\mathbf{z}|X)}{p(\mathbf{z})}\,d\mathbf{z},
\end{equation}
and for our VAE setup with Gaussian $p(\mathbf{z})$ it simplifies to
\begin{equation}
    \frac{1}{2}\sum_{i=1}^{L_d}\sum_{j=1}^{L_d}\sum_{c=1}^{L_c}\left([\mu_\theta]_{ijc}^2 + [\Sigma_\theta]_{ijc}^2 - 1 - \log[\Sigma_\theta]_{ijc}^2\right).
\end{equation}
The idea is that if the KL loss term $\beta>0$ is weighted heavily and the latent space truly converges to $p(\mathbf{z})=\mathcal{N}(\mathbf{0},I_{L_d \times L_d \times L_c})$, then the VAE can be used in a purely generative form by simply sampling from $\mathcal{N}(\mathbf{0},I_{L_d \times L_d \times L_c})$ and passing those samples through the decoder to generate random samples $\hat{X}$. If we instead use a smaller $\beta \ll 1$, then the VAE has more freedom to arrange the latent distribution in an optimal way while developing interesting latent structure. Because our approach will be to do conditional diffusion-based sampling, where we guide the diffusion generative process, we do not need to strictly force convergence to the latent prior and therefore we use $\beta=10^{-3}$ to allow more flexibility in the learned latent space while also gently pushing that latent representations to stay small as a regularization. It is worth noting that very similar results are achieved with a vanilla AE by simply adding an $L_2$ norm-based regularization on the latent encoding. 

%%%%%%%%%%%%%%%%%%%%%%%%%%%%%%%%%%%%%%%%%%%%%%%%%%%%%%%%%%%%%%%%%%
%%%%%%%%%%%%%%%%%%%%%%%%%%%%%%%%%%%%%%%%%%%%%%%%%%%%%%%%%%%%%%%%%%
\subsection{Denoising Diffusion Probabilistic Models}
%%%%%%%%%%%%%%%%%%%%%%%%%%%%%%%%%%%%%%%%%%%%%%%%%%%%%%%%%%%%%%%%%%
%%%%%%%%%%%%%%%%%%%%%%%%%%%%%%%%%%%%%%%%%%%%%%%%%%%%%%%%%%%%%%%%%%

To get an idea of how generative diffusion models work, we consider an analytically intractable data distribution $p_{\mathrm{data}}$ of high-dimensional objects $\x$ such as our latent embeddings of the 2D MHD projections with $\x\in\mathbb{R}^{d} = \mathbb{R}^{16\times 16\times 4}$, so $d=1024$. The goal of generative diffusion models is to generate new samples of $p_{\mathrm{data}}$ by transporting samples from an easy-to-sample analytical distribution $p_{\mathrm{target}}$. To do this we consider the diffusion SDE:
\begin{equation}
    \x(0) \sim p_{\mathrm{data}}, \quad \, \text{d}\x = \Bmu(\x,t)\, \text{d}t + \Bsigma(\x,t)\, \text{d}w, \label{diffSDE}
\end{equation}
where $w(t)$ is a Wiener process, $\Bmu(\x,t)$ is known as the drift term and $\Bsigma(\x,t)$  defines a diffusion tensor $\mathbf{D}(\x,t)=\Bsigma(\x,t)\,\Bsigma^T(\x,t)/2$. Consider the probability path $p(\x,t)$ driven by $\Bmu$ and $\Bsigma$ such that $p(\x,0)=p_{\mathrm{data}}(\x)$ and $p(\x,1)=p_{\mathrm{target}}(\x)$, then SDE (\ref{diffSDE}) transports samples $\x(0) \sim p_{\mathrm{data}}$ to samples $\x(1) \sim p_{\mathrm{target}}$. 

As proven in \cite{anderson1982reverse}, there exists a reverse-time SDE (\ref{diffSDE}):
\begin{eqnarray}
    \text{d}\x &=& \left [ \Bmu(\x,t) - \nabla_\x\cdot \left(\Bsigma(\x,t)\Bsigma^T(\x,t)\right) \right . \nonumber \\
   && \left . - \Bsigma(\x,t)\,\Bsigma^T(\x,t) \nabla_\x \log p\left(\x, t\right) \right ]\, \text{d}t \nonumber \\
   && + \Bsigma(\x,t) \,\text{d} w, \label{diffSDEreverse}
\end{eqnarray}
which is used to transport samples $\x(1)\sim p_{\mathrm{target}}$ for generating novel samples $\x(0)\sim p_{\mathrm{data}}$. A generative diffusion model utilizes a UNet or Diffusion Transformer $\mathbf{M}^\theta(\x,t)$ to model the reverse SDE (\ref{diffSDEreverse}) as:
\begin{equation}
    \x(1)\sim p_{\mathrm{target}}, \quad \,\text{d}\x = \mathbf{M}^\theta(\x,t)\,\text{d}t + \Bsigma(\x,t)\,\text{d}w \label{diffSDEreverse_NN}
\end{equation}
where model weights $\theta$ are parameterized by forward sample paths of Eq.~(\ref{diffSDE}) from data samples.

This is a very general class of generative models based on the diffusion SDEs, they provide a unifying framework that connects a broad spectrum of modern generative models, including normalizing flows \cite{papamakarios2021normalizing}, flow matching \cite{lipman2022flow}, consistency models \cite{song2023consistency}, posterior mean matching \cite{salazar2024posterior}, energy models \cite{lecun2006tutorial}, and optimal transport \cite{peyre2019computational}. For example, continuous-time normalizing flow (Neural ODEs) can be interpreted as diffusion processes with vanishing diffusion and learnable drift and consistency models can be viewed as alternative parametrizations of diffusion-based score learning.

Once our VAE is trained, we freeze the encoder and decoder and we generate latent representations of all of our data, which gives us our latent training dataset $D_\mathbf{z}=\{ \mathbf{z}_1, \mathbf{z}_2, \dots, \mathbf{z}_N\}$ with $\mathbf{z} \sim \rho_{\theta}(\mathbf{z})$ which is the VAE-based encoding our our entire training data set $D_X=\{X_1,X_2,\dots,X_N\}$ with $X_i \sim \rho_{\mathrm{data}}(X)$. The goal of our denoising diffusion probabilistic model (DDPM) is to learn how to generate estimates $\hat{\mathbf{z}}$ of samples $\mathbf{z} \sim \rho_{\theta}(\mathbf{z})$ in two particular forward time flow and backward time flow conditioned auto-regressive ways such that we recover samples according to
\begin{eqnarray}
    \hat{\mathbf{z}}_{t+1} &\sim& p_\theta(\mathbf{z}_{t+1} | \mathbf{z}_t, \mathbf{z}_{t-1},c_d=+1), \\
    \hat{\mathbf{z}}_{t-2} &\sim& p_\theta(\mathbf{z}_{t-2} | \mathbf{z}_t, \mathbf{z}_{t-1},c_d=-1).
\end{eqnarray}

Encoding a full MHD trajectory $\{X_t\}_{t=1}^{T}$ with the frozen VAE yields a latent trajectory $\{\mathbf{z}_t\}_{t=1}^{T}$ distributed according to a joint latent density $\rho_\theta(\mathbf{z}_1,\dots,\mathbf{z}_T)$ that inherits the temporal correlations of the underlying plasma dynamics. Rather than model this joint density directly, the diffusion model learns its local conditional factors. Because the dynamics may be unrolled in either temporal direction, we condition on an ordered pair of adjacent latent states together with a direction flag $c_d\in\{+1,-1\}$ and learn the single bidirectional transition density
\begin{equation}
\hat{\mathbf{z}}_{t+c_d}\ \sim\
p_\theta\!\left(\mathbf{z}_{t+c_d}\ \middle|\ \mathbf{z}_t,\ \mathbf{z}_{t-c_d},\ c_d\right),
\end{equation}
in which a single set of weights $\theta$ represents both temporal directions. Taking $c_d=+1$ gives the forward transition
$p_\theta(\mathbf{z}_{t+1}\mid\mathbf{z}_t,\mathbf{z}_{t-1},+1)$, which advances the system one step into the future, while $c_d=-1$ gives the backward transition $p_\theta(\mathbf{z}_{t-1}\mid\mathbf{z}_t,\mathbf{z}_{t+1},-1)$, which reconstructs the preceding state and lets the same model serve as an inverse solver. Under the second-order Markov approximation implied by conditioning on two states, these are exactly the two-sided conditional factors of $\rho_\theta(\mathbf{z}_1,\dots,\mathbf{z}_T)$ obtained by applying the chain rule in increasing or decreasing time order.

A single denoiser network realizes this density by learning to reverse a Gaussian noising of the target latent $\mathbf{z}_{t+c_d}$ while conditioned on the context $(\mathbf{z}_t,\mathbf{z}_{t-c_d},c_d)$; drawing both signs of $c_d$ from the training trajectories forces the shared weights to denoise consistently toward the future and toward the past. Longer trajectories are then generated autoregressively: starting from a seed pair of adjacent latent states, each newly sampled state is appended to the context and used to predict the next in the chosen direction, so that the joint density of a generated segment factorizes as a product of learned transitions,
\begin{eqnarray}
&& p_\theta\!\left(\mathbf{z}_{t_0+c_d},\,\mathbf{z}_{t_0+2c_d},\dots\ \middle|\
\mathbf{z}_{t_0},\,\mathbf{z}_{t_0-c_d},\,c_d\right) \nonumber \\
&& =\prod_{m\ge 1}
p_\theta\!\left(\mathbf{z}_{t_0+mc_d}\ \middle|\
\mathbf{z}_{t_0+(m-1)c_d},\,\mathbf{z}_{t_0+(m-2)c_d},\,c_d\right).
\end{eqnarray}
The associated physical fields are recovered at each step by decoding through the frozen VAE, $\hat{X}_{t+c_d}=\mathcal{D}_\theta(\hat{\mathbf{z}}_{t+c_d})$. Because the same density drives both directions, a trajectory can be integrated forward from an
initial condition as a surrogate solver, or backward from a final measured state to infer the plasma history.

\begin{figure*}[ht]
\centering
\includegraphics[width=1.0\linewidth]{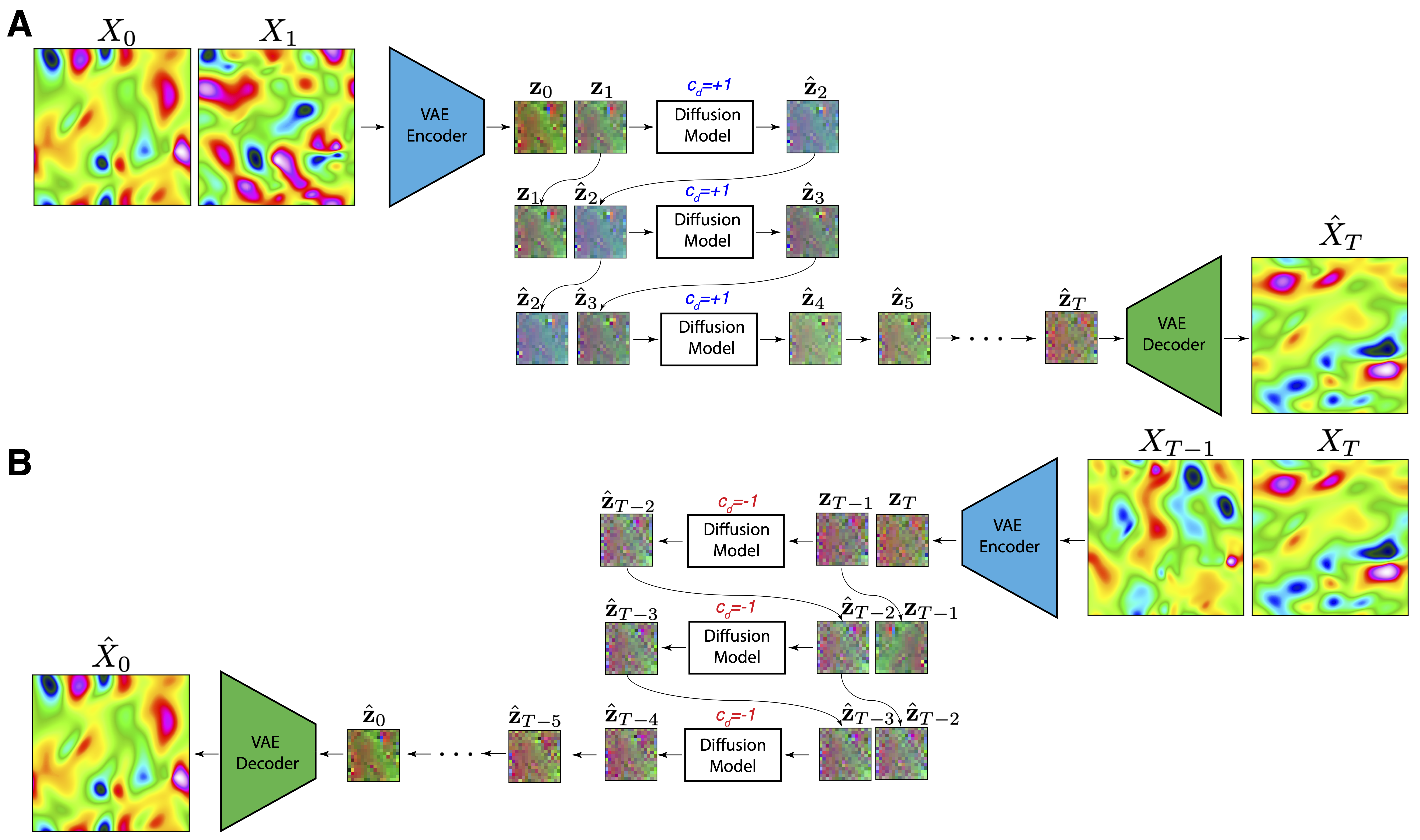}
\caption{\textbf{A}: Forward autoregressive rollout predicting MHD fields at later times based on initial conditions. \textbf{B}: Backward  autoregressive rollout solving the inverse problem of predicting previous MHD states based on later times.}
\label{fig:rollout}
\end{figure*}

We denote the diffusion (denoising) step by a superscript $k=0,\dots,K$ to keep it
distinct from the physical time index $t$, with $\mathbf{z}_{t+c_d}^{(0)}\equiv
\mathbf{z}_{t+c_d}$ the clean target and $\mathbf{z}_{t+c_d}^{(K)}\sim
\mathcal{N}(\mathbf{0},I)$ pure noise. The target is corrupted by a variance schedule
$\{\beta_k\}_{k=1}^{K}$, which in closed form reads
\begin{eqnarray}
\mathbf{z}_{t+c_d}^{(k)}&=&\sqrt{\bar\alpha_k}\,\mathbf{z}_{t+c_d}
+\sqrt{1-\bar\alpha_k}\,\boldsymbol\epsilon,\qquad
\boldsymbol\epsilon\sim\mathcal{N}(\mathbf{0},I),\nonumber \\
\alpha_k&=&1-\beta_k,\ \ \bar\alpha_k=\prod_{j=1}^{k}\alpha_j .
\end{eqnarray}
Generation runs the context-conditioned reverse process
\begin{equation}
p_\theta\!\left(\mathbf{z}_{t+c_d}^{(k-1)}\ \middle|\
\mathbf{z}_{t+c_d}^{(k)},\,\mathbf{z}_t,\,\mathbf{z}_{t-c_d},\,c_d\right)
=\mathcal{N}\!\left(\mathbf{z}_{t+c_d}^{(k-1)};\ \boldsymbol\mu_\theta,\ \sigma_k^2 I\right),
\end{equation}
where
\begin{equation}
\boldsymbol\mu_\theta=\frac{1}{\sqrt{\alpha_k}}\!\left(\mathbf{z}_{t+c_d}^{(k)}
-\frac{\beta_k}{\sqrt{1-\bar\alpha_k}}\,\boldsymbol\epsilon_\theta\right),
\end{equation}
where the denoiser $\boldsymbol\epsilon_\theta=\boldsymbol\epsilon_\theta(
\mathbf{z}_{t+c_d}^{(k)},k,\mathbf{z}_t,\mathbf{z}_{t-c_d},c_d)$ is the network that is
actually trained, by the conditional noise-prediction loss $\mathcal{L}(\theta)$ defined as
\begin{equation}
    \mathcal{L}(\theta) = \mathbb{E}_{t,c_d,k,\boldsymbol\epsilon}
\left[\left\|\,\boldsymbol\epsilon -\boldsymbol\epsilon_\theta \right\|_2^2\right], 
\end{equation}
where
\begin{equation}
\boldsymbol\epsilon_\theta = \boldsymbol\epsilon_\theta\!\left(
\sqrt{\bar\alpha_k}\,\mathbf{z}_{t+c_d}+\sqrt{1-\bar\alpha_k}\,\boldsymbol\epsilon,\
k,\ \mathbf{z}_t,\,\mathbf{z}_{t-c_d},\,c_d\right),
\end{equation}
with the physical-time index $t$ and direction $c_d$ drawn uniformly over the training trajectories and $k\sim\mathcal{U}\{1,\dots,K\}$. Marginalizing the reverse chain over the intermediate noised states $\mathbf{z}_{t+c_d}^{(1:K)}$ returns the transition density quoted above. The forward vs backward autoregressive rollout setup is shown in Figure \ref{fig:rollout}.

\begin{figure}[ht]
\centering
\includegraphics[width=1.0\linewidth]{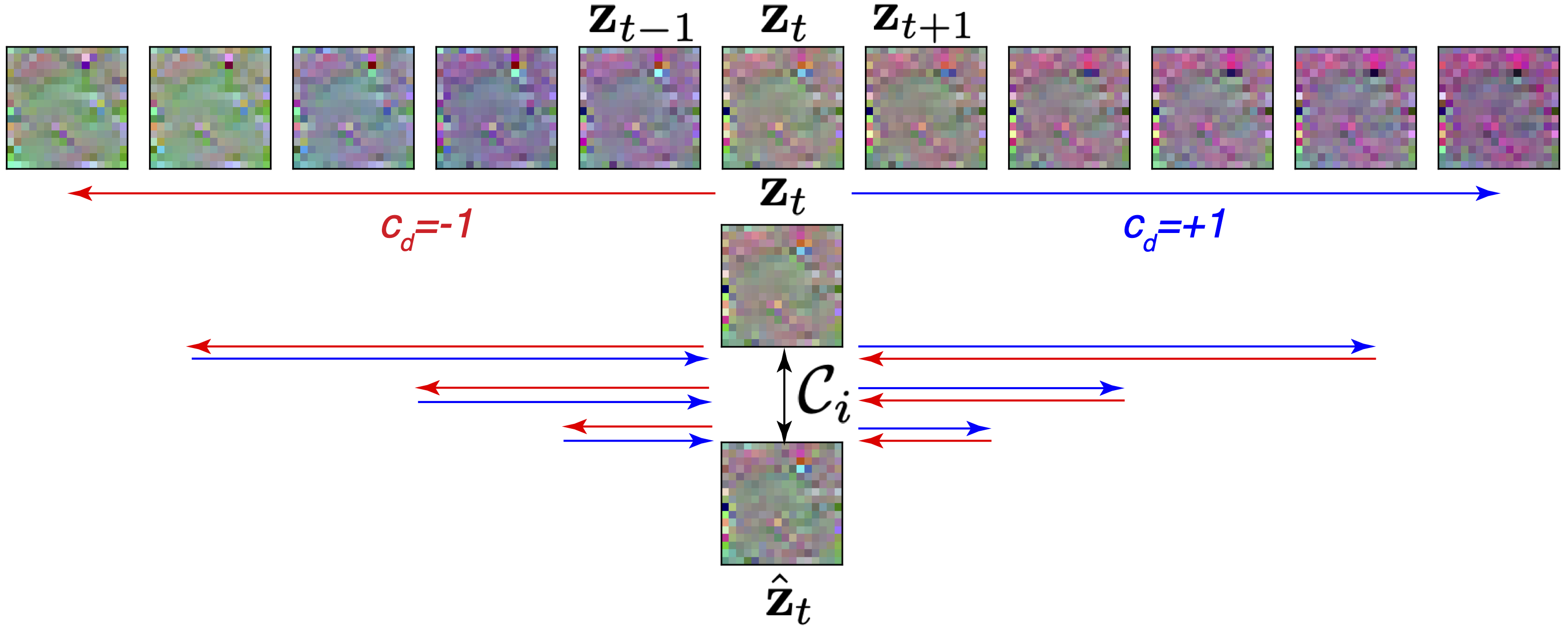}
\caption{The idea of consistency cycling for self-supervised error detection is shown. As the forward and backward cycles in either direction get longer, we expect to return to a starting condition with increasing error.}
\label{fig:consistency}
\end{figure}

%%%%%%%%%%%%%%%%%%%%%%%%%%%%%%%%%%%%%%%%%%%%%%%%%%%%%%%%%%%%%%%%%%
%%%%%%%%%%%%%%%%%%%%%%%%%%%%%%%%%%%%%%%%%%%%%%%%%%%%%%%%%%%%%%%%%%
%%%%%%%%%%%%%%%%%%%%%%%%%%%%%%%%%%%%%%%%%%%%%%%%%%%%%%%%%%%%%%%%%%
%%%%%%%%%%%%%%%%%%%%%%%%%%%%%%%%%%%%%%%%%%%%%%%%%%%%%%%%%%%%%%%%%%
\section{Self-supervised Consistency Error Detection}
%%%%%%%%%%%%%%%%%%%%%%%%%%%%%%%%%%%%%%%%%%%%%%%%%%%%%%%%%%%%%%%%%%
%%%%%%%%%%%%%%%%%%%%%%%%%%%%%%%%%%%%%%%%%%%%%%%%%%%%%%%%%%%%%%%%%%
%%%%%%%%%%%%%%%%%%%%%%%%%%%%%%%%%%%%%%%%%%%%%%%%%%%%%%%%%%%%%%%%%%
%%%%%%%%%%%%%%%%%%%%%%%%%%%%%%%%%%%%%%%%%%%%%%%%%%%%%%%%%%%%%%%%%%

When the model is deployed, it is rolled out autoregressively from a known seed
pair into a regime where no ground-truth latents are available. In this setting the
quantity we actually care about, the forward rollout error at depth $i$,
\begin{equation}
\mathcal{E}_i \;=\; \mathrm{MSE}\!\left(\mathbf{z}_{t+i},\,\hat{\mathbf{z}}_{t+i}\right),
\
\mathrm{MSE}(\mathbf{a},\mathbf{b})=\frac{1}{L_d^2 L_c}\sum_{p,q,c}\bigl(a_{pqc}-b_{pqc}\bigr)^2,
\end{equation}
cannot be measured, because $\mathbf{z}_{t+i}$ is precisely the unknown state the
model is predicting. Since each learned transition is only approximate and these
approximations compound from step to step, $\mathcal{E}_i$ grows with rollout depth,
and we would like a per-trajectory estimate of how far we can trust a given rollout
\emph{before} committing to it. The bidirectionality of the model supplies such an
estimate at no additional training cost.

The key observation is that, because a single network represents both temporal
directions, any forward rollout can be reversed by a backward rollout of equal
length, and an error-free model must return exactly to where it started. Concretely,
we roll forward $i$ steps from the true seed pair,
\begin{equation}
\hat{\mathbf{z}}_{t+m}\;\sim\;p_\theta\!\left(\,\cdot\;\middle|\;
\hat{\mathbf{z}}_{t+m-1},\,\hat{\mathbf{z}}_{t+m-2},\,c_d=+1\right),
\qquad m=1,\dots,i,
\end{equation}
with the seed convention $\hat{\mathbf{z}}_t\equiv\mathbf{z}_t$,
$\hat{\mathbf{z}}_{t-1}\equiv\mathbf{z}_{t-1}$. We then flip the direction flag and,
seeding the return pass with the terminal pair
$(\hat{\mathbf{z}}_{t+i-1},\hat{\mathbf{z}}_{t+i})$, roll backward the same $i$ steps,
\begin{equation}
\tilde{\mathbf{z}}^{(i)}_{t+i-1-m}\;\sim\;p_\theta\!\left(\,\cdot\;\middle|\;
\tilde{\mathbf{z}}^{(i)}_{t+i-m},\,\tilde{\mathbf{z}}^{(i)}_{t+i-m+1},\,c_d=-1\right),
\ m=1,\dots,i,
\end{equation}
where the superscript $(i)$ records the depth at which the trajectory was turned
around. The composition of the two passes is a closed cycle that maps the seed pair
back onto itself,
\begin{eqnarray}
&& (\mathbf{z}_{t-1},\mathbf{z}_t)
\;\xrightarrow{\;\Phi^{i}_{+}\;}\;
(\hat{\mathbf{z}}_{t+i-1},\hat{\mathbf{z}}_{t+i})
\;\xrightarrow{\;\Phi^{i}_{-}\;}\;
(\tilde{\mathbf{z}}^{(i)}_{t-1},\,\tilde{\mathbf{z}}^{(i)}_{t}), \\
&& \Phi^{i}_{-}\!\circ\Phi^{i}_{+}\stackrel{\text{ideal}}{=}\mathrm{Id},
\end{eqnarray}
and we define the self-supervised consistency error at depth $i$ as the round-trip
discrepancy on the two recovered seed latents,
\begin{equation}
\mathcal{C}_i \;=\; \tfrac{1}{2}\Bigl[\,
\mathrm{MSE}\!\left(\mathbf{z}_{t-1},\,\tilde{\mathbf{z}}^{(i)}_{t-1}\right)
+\mathrm{MSE}\!\left(\mathbf{z}_{t},\,\tilde{\mathbf{z}}^{(i)}_{t}\right)\Bigr].
\end{equation}
Crucially, every quantity entering $\mathcal{C}_i$ is available at test time: the
anchor pair $(\mathbf{z}_{t-1},\mathbf{z}_t)$ is the encoding of measured data, and
$(\tilde{\mathbf{z}}^{(i)}_{t-1},\tilde{\mathbf{z}}^{(i)}_{t})$ is produced entirely by
the model. No ground truth for any of the intermediate rolled-out states
$\hat{\mathbf{z}}_{t+1},\dots,\hat{\mathbf{z}}_{t+i}$ is ever required.

The diagnostic value of $\mathcal{C}_i$ follows from a simple consistency argument. If
the forward and backward maps were mutually exact, the cycle would be the identity
and $\mathcal{C}_i=0$ for every $i$; any departure from zero therefore certifies that
error has entered somewhere along the $2i$-step round trip. Cycle consistency is thus
a \emph{necessary} condition for an accurate rollout, and while it is not strictly
sufficient, in that compensating forward and backward errors could in principle
cancel, we find empirically that $\mathcal{C}_i$ increases monotonically with depth in
lockstep with the unobservable $\mathcal{E}_i$, so that the measurable quantity
$\mathcal{C}_i$ serves as a faithful surrogate for the error we cannot see. Because
each diffusion sample is stochastic, we make the cycle well-defined by either using
the deterministic probability-flow sampler or by averaging over $S$ independent
rollouts, $\bar{\mathcal{C}}_i=\tfrac{1}{S}\sum_{s}\mathcal{C}_i^{(s)}$, in which case
the spread across samples provides an additional, complementary uncertainty estimate.
In practice this yields a direct trust criterion: given a tolerance $\tau$, a rollout
is trusted out to the largest depth for which $\mathcal{C}_i\le\tau$, and $\mathcal{C}_i$
can equally be used to weight, branch, or terminate trajectories. The same signal
drives the adaptive feedback scheme of the next section. The anchor need not be the
original seed either; the turnaround can be performed at any point along a long
rollout and checked against any earlier latent pair that is trusted, and if a
diagnostic in physical units is preferred the discrepancy can be evaluated after
decoding, $\mathrm{MSE}(\mathcal{D}_\theta(\mathbf{z}_t),
\mathcal{D}_\theta(\tilde{\mathbf{z}}^{(i)}_{t}))$, at the cost of one extra decode. A high level overview of this consistency cylcing approach is shown in Figure \ref{fig:consistency}.

%%%%%%%%%%%%%%%%%%%%%%%%%%%%%%%%%%%%%%%%%%%%%%%%%%%%%%%%%%%%%%%%%%
%%%%%%%%%%%%%%%%%%%%%%%%%%%%%%%%%%%%%%%%%%%%%%%%%%%%%%%%%%%%%%%%%%
%%%%%%%%%%%%%%%%%%%%%%%%%%%%%%%%%%%%%%%%%%%%%%%%%%%%%%%%%%%%%%%%%%
%%%%%%%%%%%%%%%%%%%%%%%%%%%%%%%%%%%%%%%%%%%%%%%%%%%%%%%%%%%%%%%%%%
\section{Adaptive Predictions From Limited Views}
%%%%%%%%%%%%%%%%%%%%%%%%%%%%%%%%%%%%%%%%%%%%%%%%%%%%%%%%%%%%%%%%%%
%%%%%%%%%%%%%%%%%%%%%%%%%%%%%%%%%%%%%%%%%%%%%%%%%%%%%%%%%%%%%%%%%%
%%%%%%%%%%%%%%%%%%%%%%%%%%%%%%%%%%%%%%%%%%%%%%%%%%%%%%%%%%%%%%%%%%
%%%%%%%%%%%%%%%%%%%%%%%%%%%%%%%%%%%%%%%%%%%%%%%%%%%%%%%%%%%%%%%%%%

We consider a case in which we have rolled-out predictions 
\begin{equation} 
\mathbf{z}_0, \ \mathbf{z}_1, \ \hat{\mathbf{z}}_2, \dots \hat{\mathbf{z}}_{T},
\end{equation}
and we have access to a measurement of a limited number of fields at $t=T$. For example maybe we are able to measure just the density field $\rho(x,y,T)$, and we would like to use that measurement to improve our predictions of the fields which we do not have access to:
\begin{equation}
    v_x(x,y,T), \ v_y(x,y,T), \ P(x,y,T), \ B_x(x,y,T), \ B_y(x,y,T).
\end{equation}
For this case we consider the application of a model-independent adaptive feedback approach known as bounded extremum seeking (ES). Bounded ES is applicable to dynamic systems of the form
\begin{equation}
    \dot{\mathbf{x}} = \mathbf{f}(\mathbf{x},t) + \mathbf{g}(\mathbf{x},t)\mathbf{u}(\mathbf{x},t),
\end{equation}
where $\mathbf{f}$ and $\mathbf{g}$ are unknown and time-varying functions, and the goal is to minimize an analytically unknown cost function $C(\mathbf{x},t)$. Bounded ES was invented in \cite{scheinker2013model}, it provides guaranteed bounds on control efforts and parameter update rates despite acting on noisy, analytically unknown time-varying systems, and has been studied for a wide range of systems including ones with non-differentiable and discontinuous controllers \cite{scheinker2016bounded}. The bounded ES method has also been combined with generative multi-modal diffusion models to make them more robust for time-varying particle accelerator applications \cite{scheinker2024cdvae}, it has been applied in-hardware in particle accelerators for optimizing the rise-time of high-voltage converter modulators \cite{scheinker2013extremum} and for RF cavity tuning \cite{scheinker2014hardware}, and it has been studied for Tokamak stabilization \cite{de2022event}. Bounded ES has been combined with deep reinforcement learning (DRL) for time-varying systems and showed that bounded ES complements the learned RL policy by maintaining robustness as the system departs from the training regime, while DRL supplies rapid control when conditions remain near those seen in training \cite{saxena2025improved}, which is especially useful for robotic control in time-varying scenarios \cite{saxena2026deep}. A method has also been created for automatic Mahalanobis distance-guided switching beetween ES and the DRL controller when an out of distribution state is detected \cite{saxena2026mahalanobis}.

In this case, our cost function is defined as
\begin{eqnarray}
    C(\hat{\mathbf{z}}_T,T) &=& \left \| \rho(x,y,T) - \hat{\rho}(x,y,T) \right \|^2_1 \nonumber \\
    &=& \iint_{x,y} \left | \rho(x,y,T) - \hat{\rho}(x,y,T) \right | dx dy \nonumber \\
    &=& \iint_{x,y} \left | \rho(x,y,T) - D_\theta(\hat{\mathbf{z}}_{T,\rho}, \mathbf{c}_\rho ) \right | dx dy.
\end{eqnarray}
which we can minimize by using the model-independent bounded extremum seeking approach of \cite{scheinker2013model,scheinker2016bounded,scheinker2024cdvae}. We consider a class of systems of the form
\begin{equation}
	\dot{\mathbf{p}} = \mathbf{f}(\mathbf{p},t) + \mathbf{g}(\mathbf{p},t)\mathbf{u}(C(\mathbf{p},t)), \label{dpES}
\end{equation}
where $\mathbf{p}\in \mathbb{R}^n$  is a set of adjustable parameters, $\mathbf{f}:\mathbb{R}^n\times \mathbb{R} \rightarrow \mathbb{R}^n$ and $\mathbf{g}:\mathbb{R}^n \times \mathbb{R} \rightarrow \mathbb{R}^{n\times n}$ are analytically unknown time-varying functions, $C(\mathbf{p},t):\mathbb{R}^n\times \mathbb{R} \rightarrow \mathbb{R}$ is an analytically unknown time-varying function which we hope to minimize or maximize, and $\mathbf{u}:\mathbb{R}\times \mathbb{R}\rightarrow \mathbb{R}^n$ is our control vector. Then based on the results of \cite{scheinker2013model,scheinker2016bounded,scheinker2024cdvae}, if we choose our controller $\mathbf{u}$ such that
\begin{equation}
	u_i = \sqrt{\alpha\omega_i}\cos(\omega_i t + k C(\mathbf{p},t)), 
\end{equation}
where $\omega_i = \omega r_i$ with $r_i \neq r_j$ for all $i \neq j$, then in the limit $\omega \rightarrow \infty$, the dynamics (\ref{dpES}), will on average be given by
\begin{equation}
	\dot{\bar{\mathbf{p}}} = \mathbf{f}(\bar{\mathbf{p}},t) -\frac{k\alpha}{2} \mathbf{g}(\mathbf{p},t)^T\mathbf{g}(\mathbf{p},t)\nabla_{\mathbf{p}}C(\bar{\mathbf{p}},t), \label{dpES_ave}
\end{equation}
where $\mathbf{g}(\mathbf{p},t)^T\mathbf{g}(\mathbf{p},t) \geq 0$ ensures that we can overpower $\mathbf{f}(\mathbf{p},t)$ and drive $\mathbf{p}(t)$ towards a minimizer of $C(\mathbf{p},t)$ by choosing sufficiently large $k\alpha>0$.

%%%%%%%%%%%%%%%%%%%%%%%%%%%%%%%%%%%%%%%%%%%%%%%%%%%%%%%%%%%%%%%%%%
%%%%%%%%%%%%%%%%%%%%%%%%%%%%%%%%%%%%%%%%%%%%%%%%%%%%%%%%%%%%%%%%%%
%%%%%%%%%%%%%%%%%%%%%%%%%%%%%%%%%%%%%%%%%%%%%%%%%%%%%%%%%%%%%%%%%%
%%%%%%%%%%%%%%%%%%%%%%%%%%%%%%%%%%%%%%%%%%%%%%%%%%%%%%%%%%%%%%%%%%
\section{VAE Reconstructions}
\begin{figure}[ht]
\centering
\includegraphics[width=1.0\linewidth]{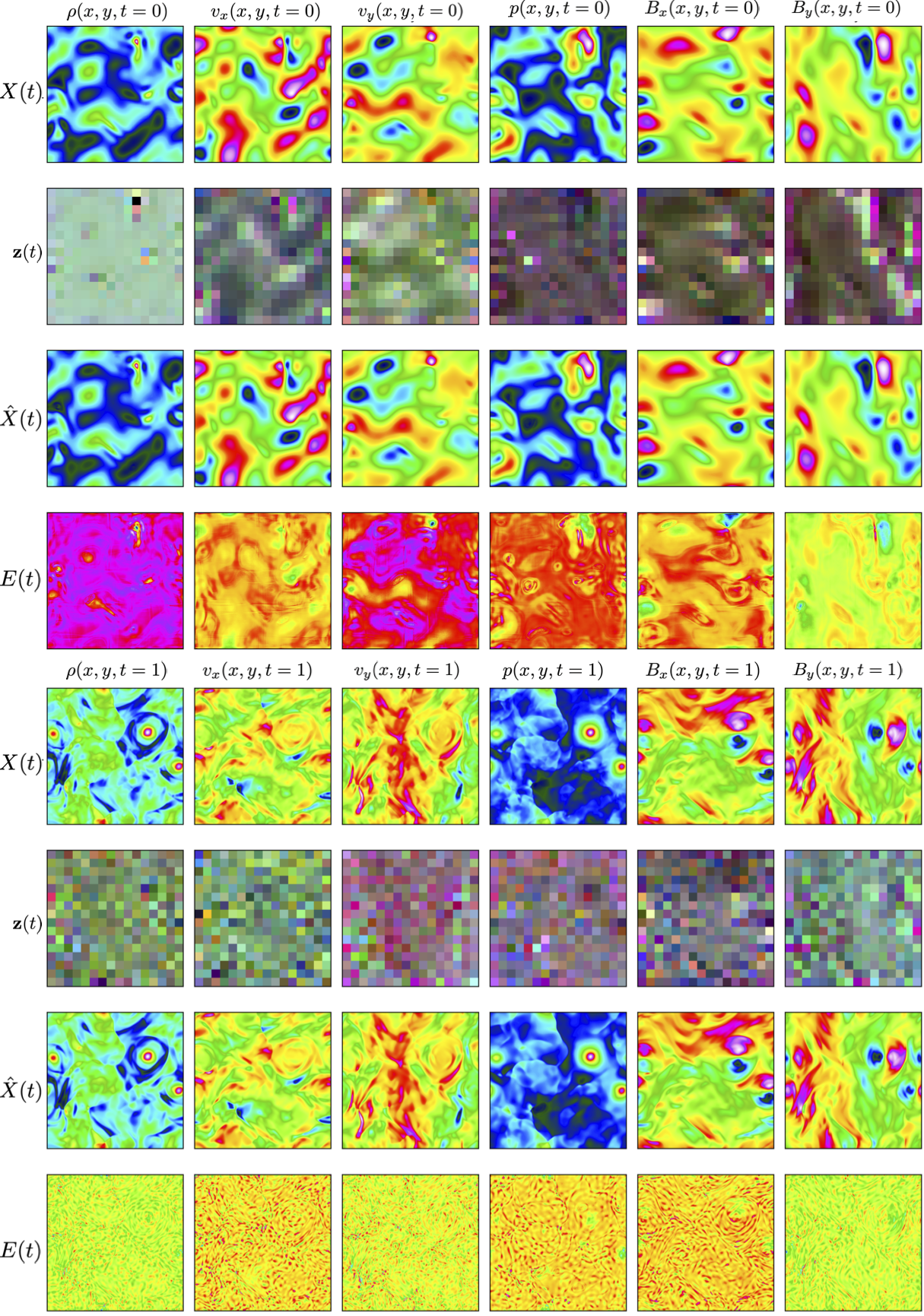}
\caption{True MHD fields are shown along with their latent embeddings and reconstructions, as well as image-wise errors for one simulation at time steps $t=0$ and $t=1$.}
\label{fig:VAE_PREDS}
\end{figure}

%%%%%%%%%%%%%%%%%%%%%%%%%%%%%%%%%%%%%%%%%%%%%%%%%%%%%%%%%%%%%%%%%%
%%%%%%%%%%%%%%%%%%%%%%%%%%%%%%%%%%%%%%%%%%%%%%%%%%%%%%%%%%%%%%%%%%
%%%%%%%%%%%%%%%%%%%%%%%%%%%%%%%%%%%%%%%%%%%%%%%%%%%%%%%%%%%%%%%%%%
%%%%%%%%%%%%%%%%%%%%%%%%%%%%%%%%%%%%%%%%%%%%%%%%%%%%%%%%%%%%%%%%%%
As a first step, we need to train the VAE and confirm that it can compress and accurately reconstruct the MHD fields because the accuracy of the overall final predictions, regardless of how the latent space dynamics are modeled, will be bottle-necked by the decoder's ability to create accurate reconstructions. One example of MHD fields along with their latent embeddings and reconstructions, as well as image-wise errors for one simulation at time steps $t=0$ and $t=1$ is shown in Figure \ref{fig:VAE_PREDS}.

\begin{figure*}[ht]
\centering
\includegraphics[width=1.0\linewidth]{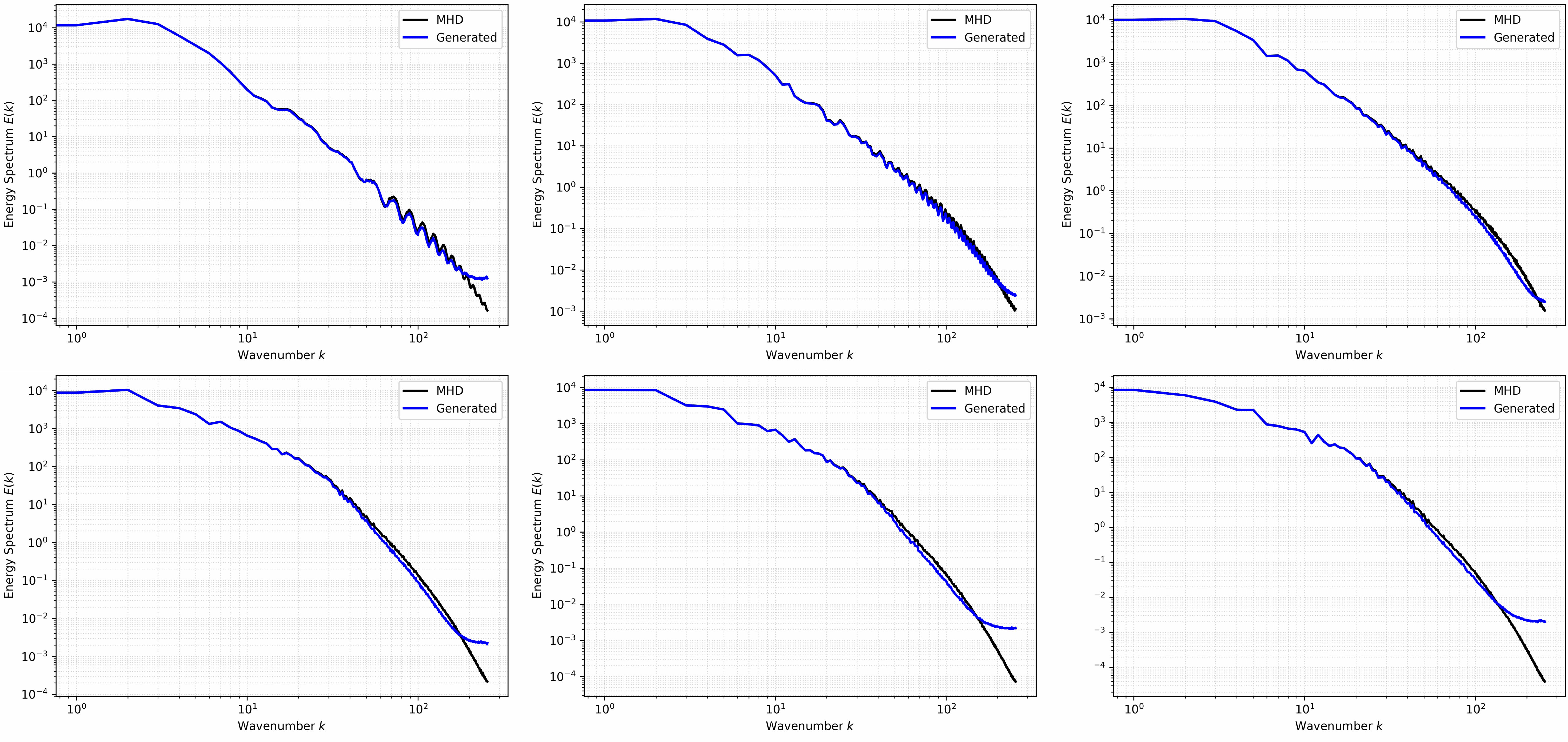}
\caption{Predicted vs true energy spectra are shown for 6 randomly chosen examples from the validation data.}
\label{fig:spectra}
\end{figure*}

By eye, the reconstructions look good and have very low mean squared error, but the more important quantitative test is to look at the power spectrum of the predictions versus the true fields to quantify how well high frequency low-level details are captured. In order to calculate the energy spectrum we first compute the Fourier transforms
\begin{equation}
    \hat{f}(\mathbf{k}) = \sum_{\mathbf{x}} f(\mathbf{x})\, e^{-i 2\pi\, \mathbf{k}\cdot\mathbf{x}/N},
\quad f \in \{v_x, v_y, B_x, B_y\},
\end{equation}
the energy spectral density is then calculated as
\begin{equation}
    E_{2\mathrm{D}}(\mathbf{k}) = \frac{1}{2N^2}\left( |\hat{v}_x(\mathbf{k})|^2 + |\hat{v}_y(\mathbf{k})|^2 + |\hat{B}_x(\mathbf{k})|^2 + |\hat{B}_y(\mathbf{k})|^2 \right),
\end{equation}
where
\begin{equation}
    \sum_{\mathbf{k}} E_{2\mathrm{D}}(\mathbf{k}) = \frac{1}{2}\sum_{\mathbf{x}}\left( v_x^2 + v_y^2 + B_x^2 + B_y^2 \right).
\end{equation}
We then perform radial binning
\begin{equation}
    E(k) = \sum_{\,k \,\le\, |\mathbf{k}| \,<\, k+1} E_{2\mathrm{D}}(\mathbf{k}),
\quad k = 0, 1, \dots, \tfrac{N}{2}-1,
\end{equation}
where
\begin{equation}
    k =\|\mathbf{k}\|^2=\sqrt{k_x^2 + k_y^2},
\end{equation}
and plot the results up to $k_\mathrm{max}=N/2$, truncating to the Nyquist wavenumber cutoff. Figure \ref{fig:spectra} shows true vs predicted power spectra for 6 randomly chosen time steps from the validation data set, which are seen to have good agreement to high wavenumber.

\begin{figure*}[ht]
\centering
\includegraphics[width=0.49\linewidth]{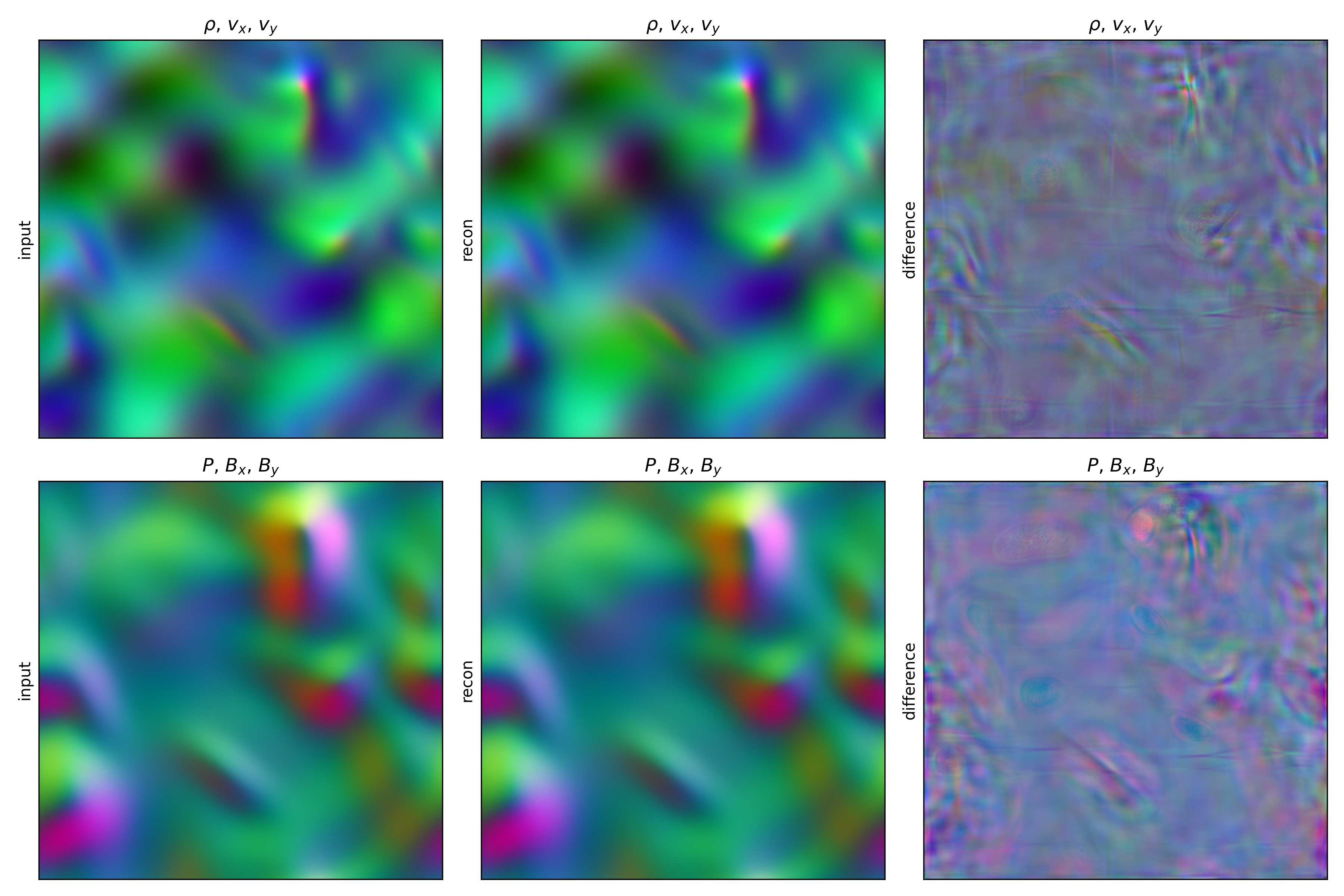}
\includegraphics[width=0.49\linewidth]{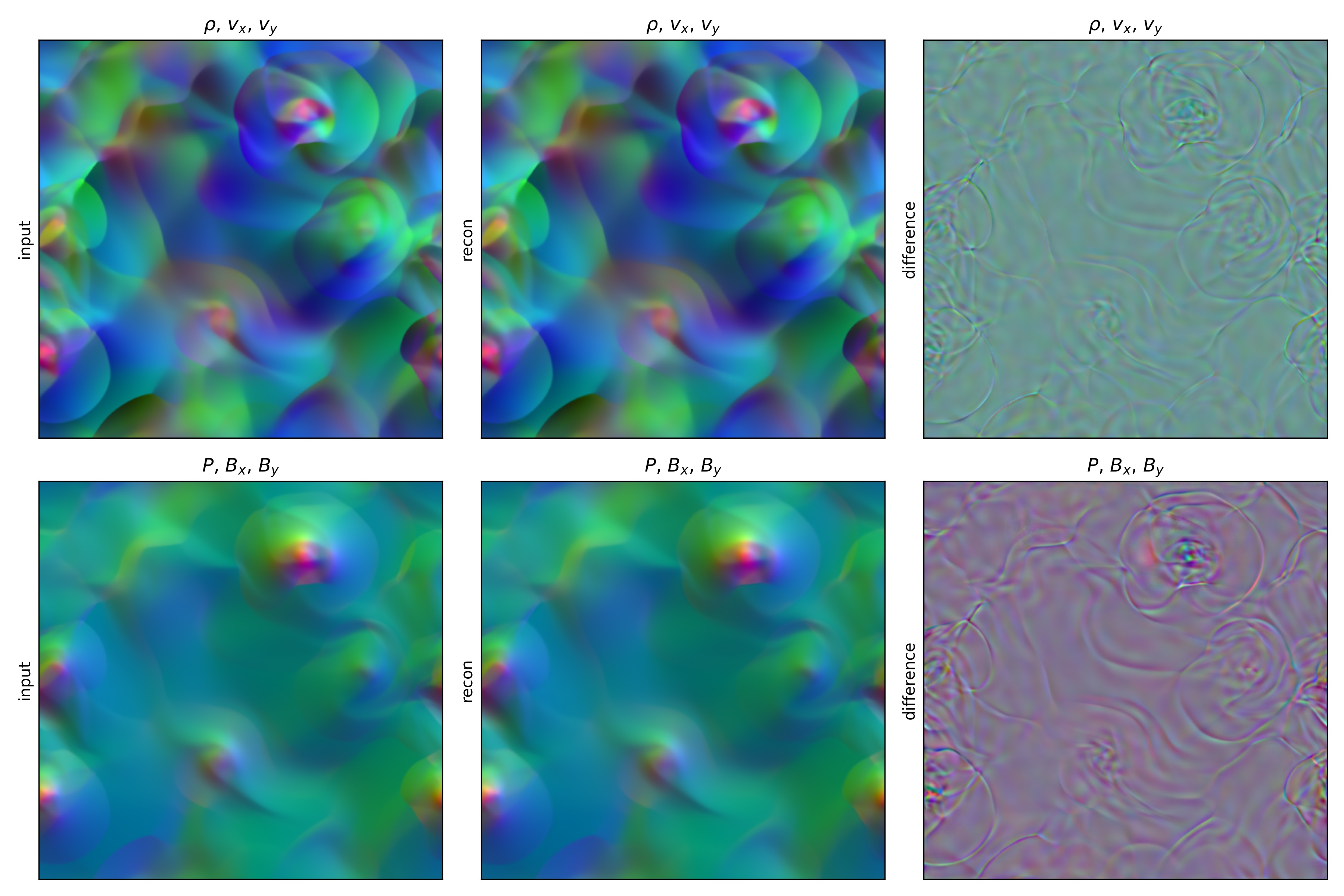}
\includegraphics[width=0.49\linewidth]{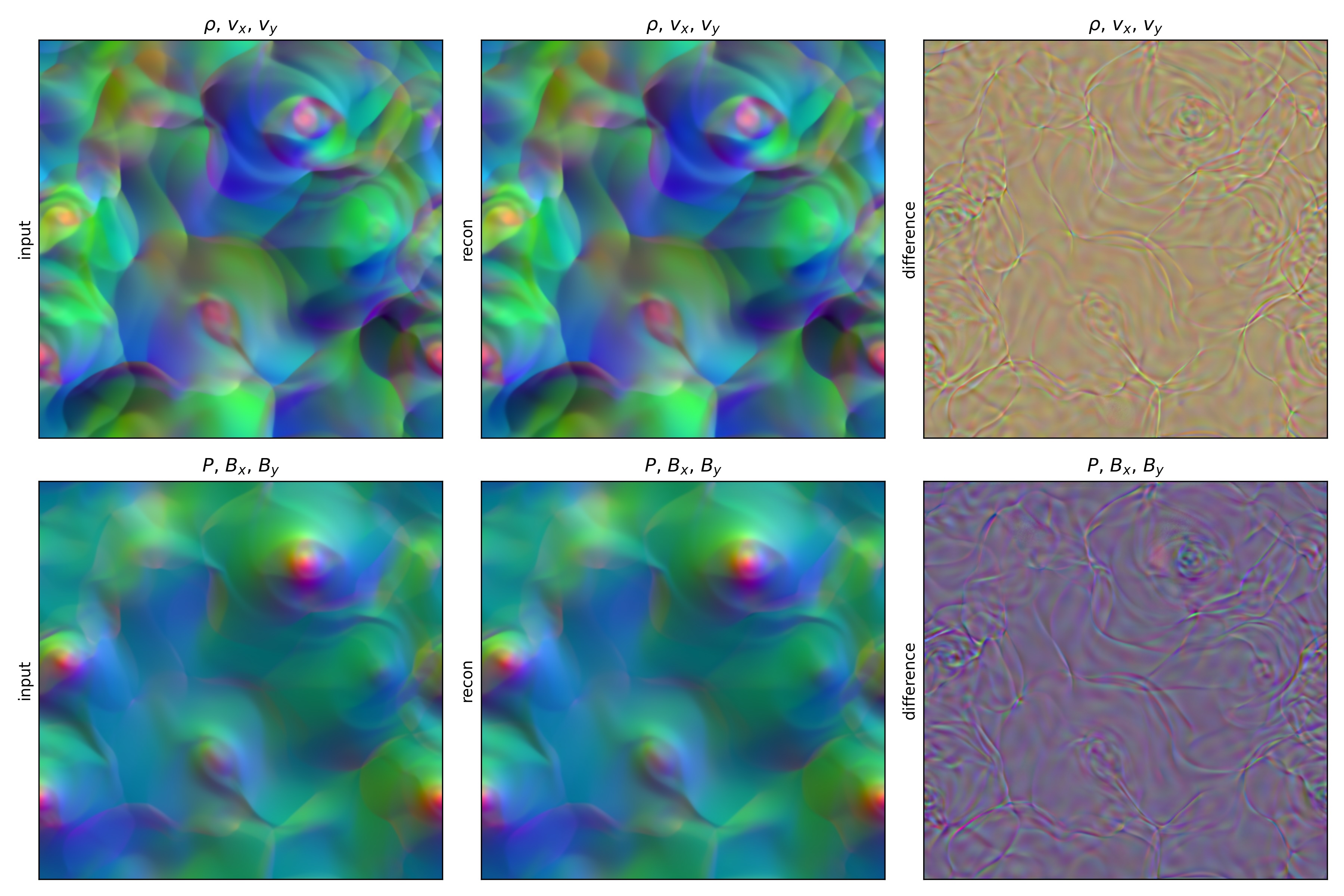}
\includegraphics[width=0.49\linewidth]{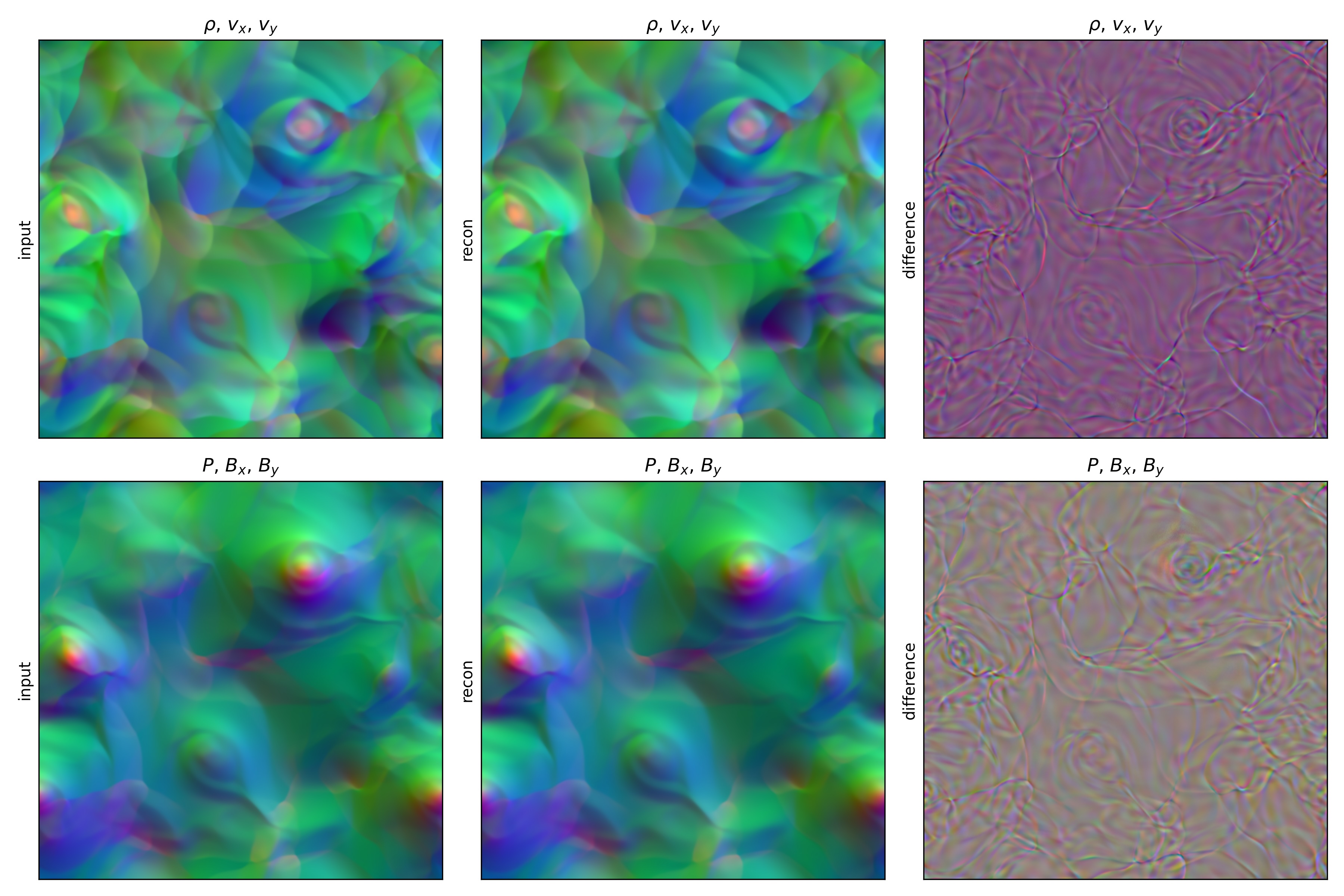}
\includegraphics[width=0.49\linewidth]{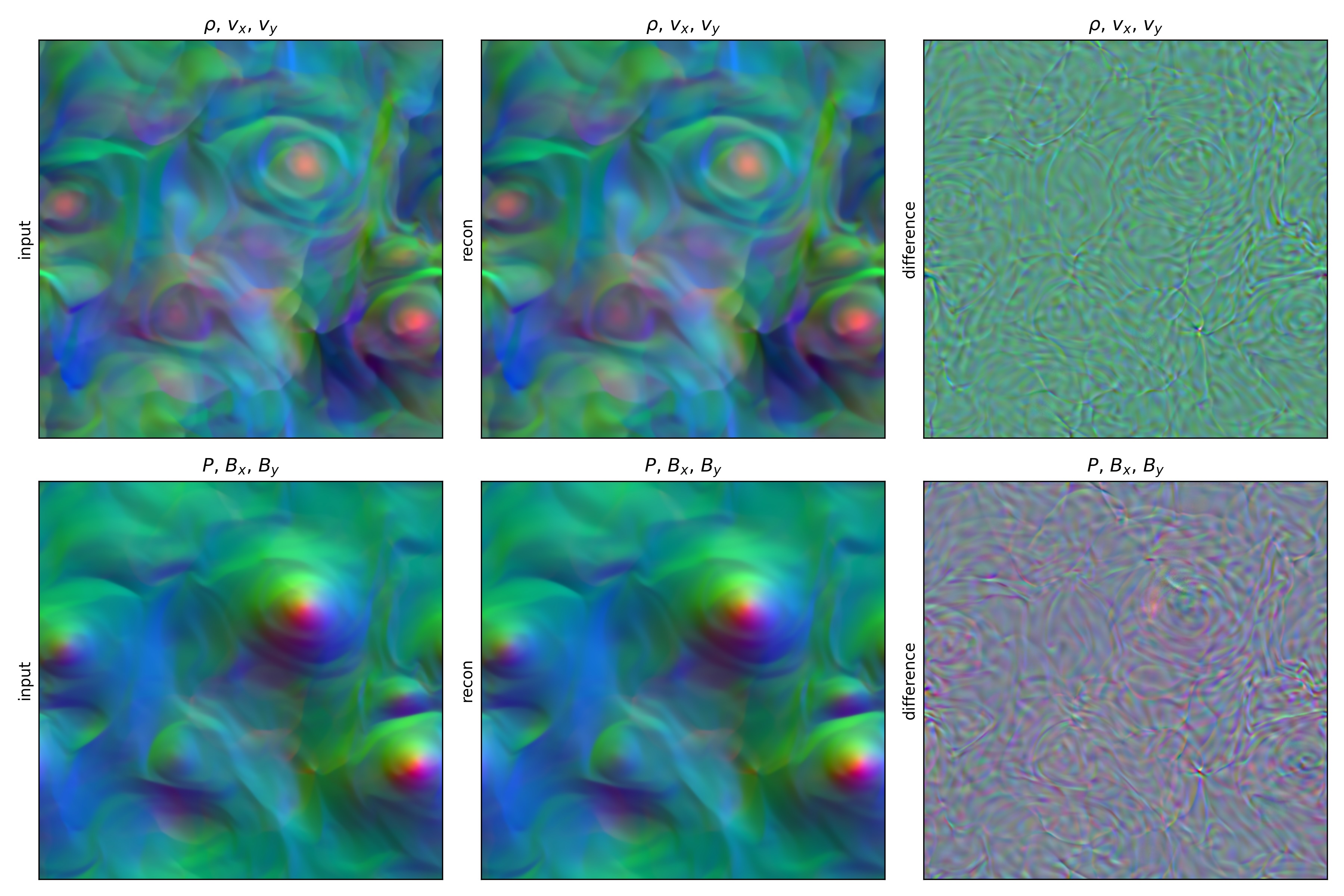}
\includegraphics[width=0.49\linewidth]{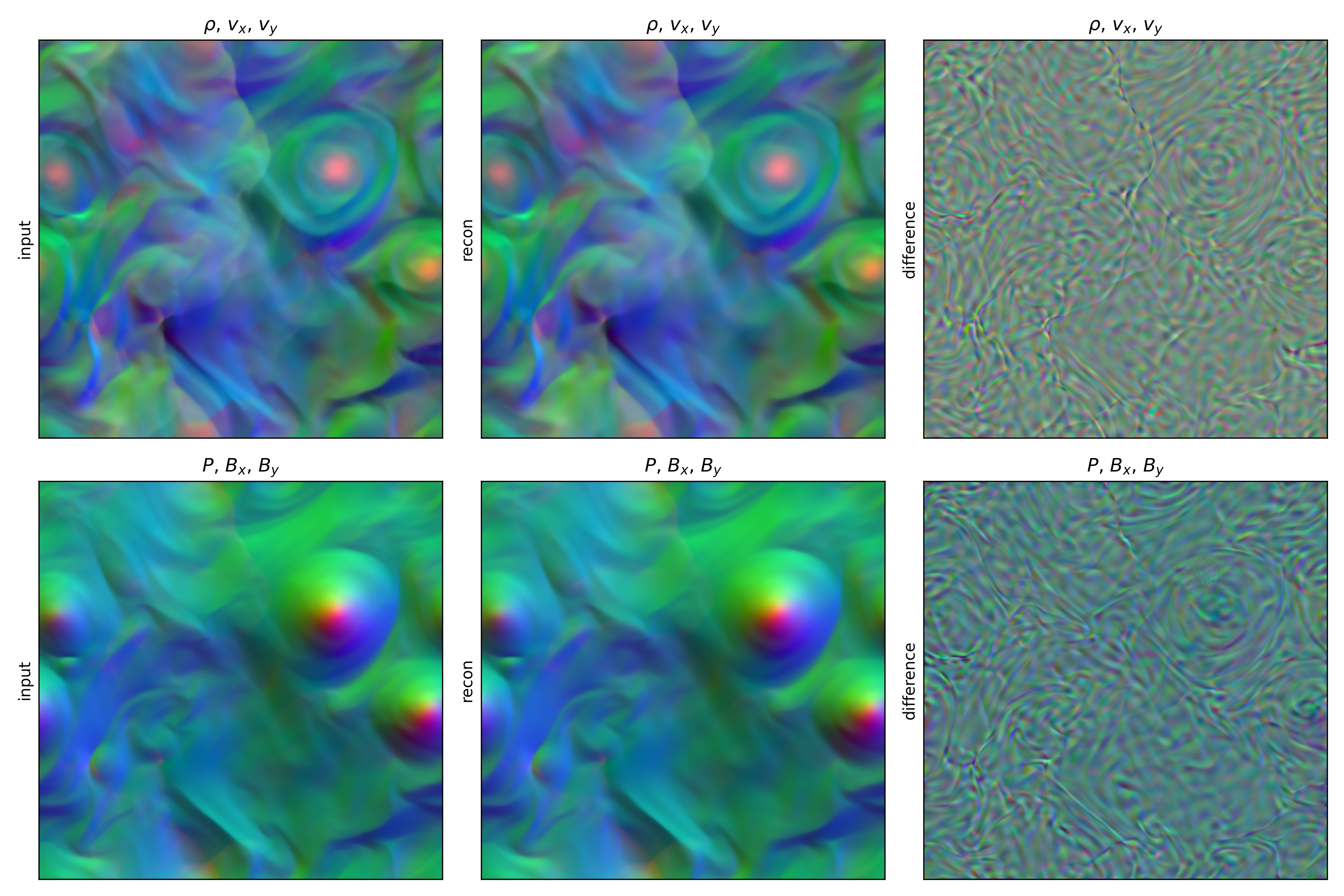}
\caption{Predicted and true fields compared at several time-steps, summarized by combining groups of three MHD fields as the RGB channels of individual images.}
\label{fig:MHD_RGB}
\end{figure*}

Another way to visualize the true vs predicted data, and to see that only very fine-grained details are missed, is to stack several 2D MHD fields together as the RGB channels of single color images, as shown in Figure \ref{fig:MHD_RGB}.

%%%%%%%%%%%%%%%%%%%%%%%%%%%%%%%%%%%%%%%%%%%%%%%%%%%%%%%%%%%%%%%%%%
%%%%%%%%%%%%%%%%%%%%%%%%%%%%%%%%%%%%%%%%%%%%%%%%%%%%%%%%%%%%%%%%%%
%%%%%%%%%%%%%%%%%%%%%%%%%%%%%%%%%%%%%%%%%%%%%%%%%%%%%%%%%%%%%%%%%%
%%%%%%%%%%%%%%%%%%%%%%%%%%%%%%%%%%%%%%%%%%%%%%%%%%%%%%%%%%%%%%%%%%
\section{Generative Diffusion}
%%%%%%%%%%%%%%%%%%%%%%%%%%%%%%%%%%%%%%%%%%%%%%%%%%%%%%%%%%%%%%%%%%
%%%%%%%%%%%%%%%%%%%%%%%%%%%%%%%%%%%%%%%%%%%%%%%%%%%%%%%%%%%%%%%%%%
%%%%%%%%%%%%%%%%%%%%%%%%%%%%%%%%%%%%%%%%%%%%%%%%%%%%%%%%%%%%%%%%%%
%%%%%%%%%%%%%%%%%%%%%%%%%%%%%%%%%%%%%%%%%%%%%%%%%%%%%%%%%%%%%%%%%%

\begin{figure}[ht]
\centering
\includegraphics[width=0.65\linewidth]{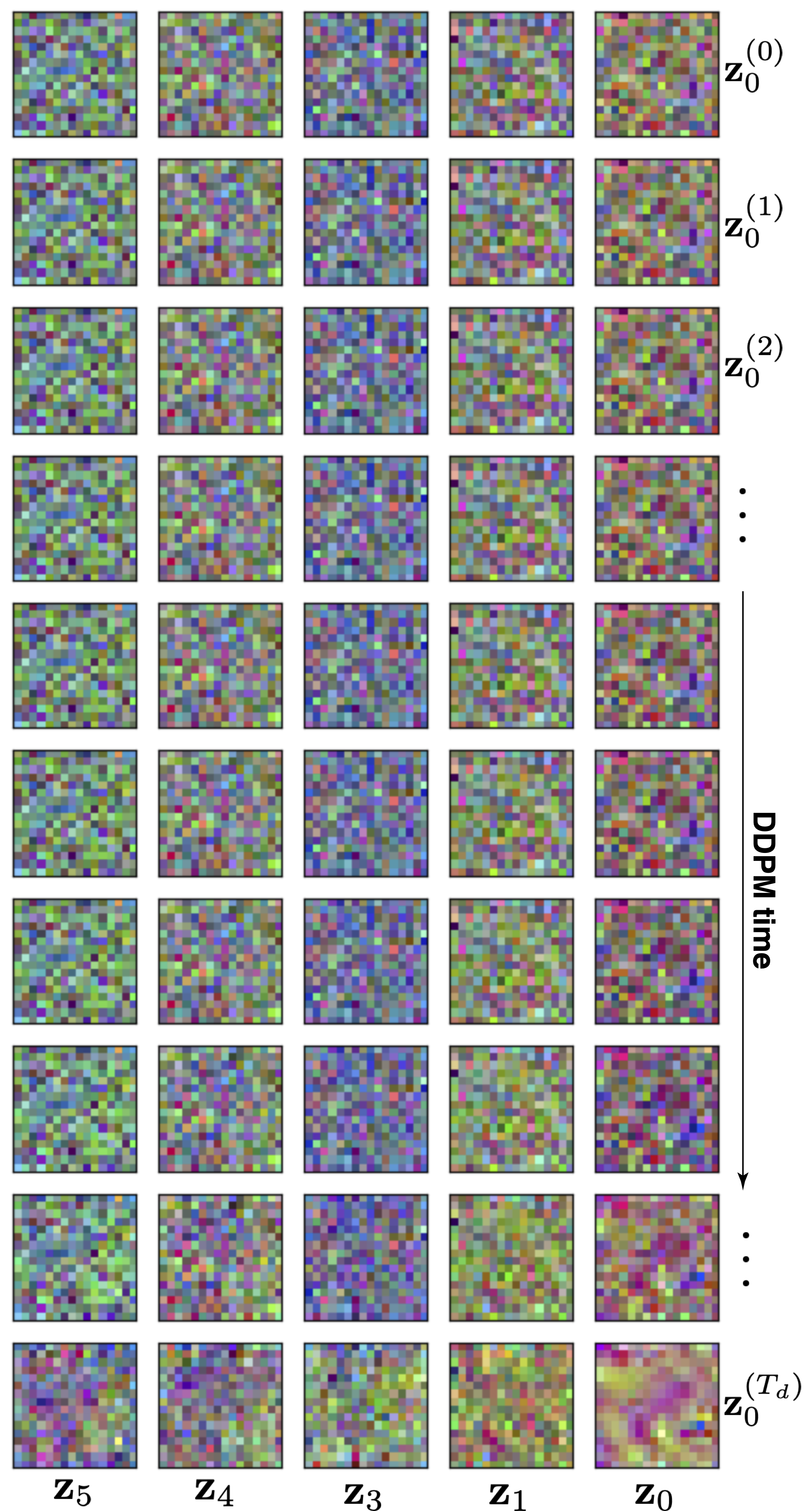}
\caption{The denoising generative process for several individual time-step predictions of $\mathbf{z}_t$ shown at various denoising diffusion steps $\mathbf{z}^{(k)}_{t}$.}
\label{fig:diff_time}
\end{figure}

At any one time-step, we build up the latent images by our iterative denoising generative process which is applying our diffusion model as an iterative denoising diffusion probabilistic model (DDPM), as shown in Figure \ref{fig:diff_time}. For the diffusion model, we utilize both a UNet-based approach as well as vision transformer (ViT)-based diffusion models. We find that the UNet trains faster, on less data, but that the ViT reaches the same and sometimes slightly better accuracy as training goes on longer and larger datasets are used.

\begin{figure*}[ht]
\centering
\includegraphics[width=1.0\linewidth]{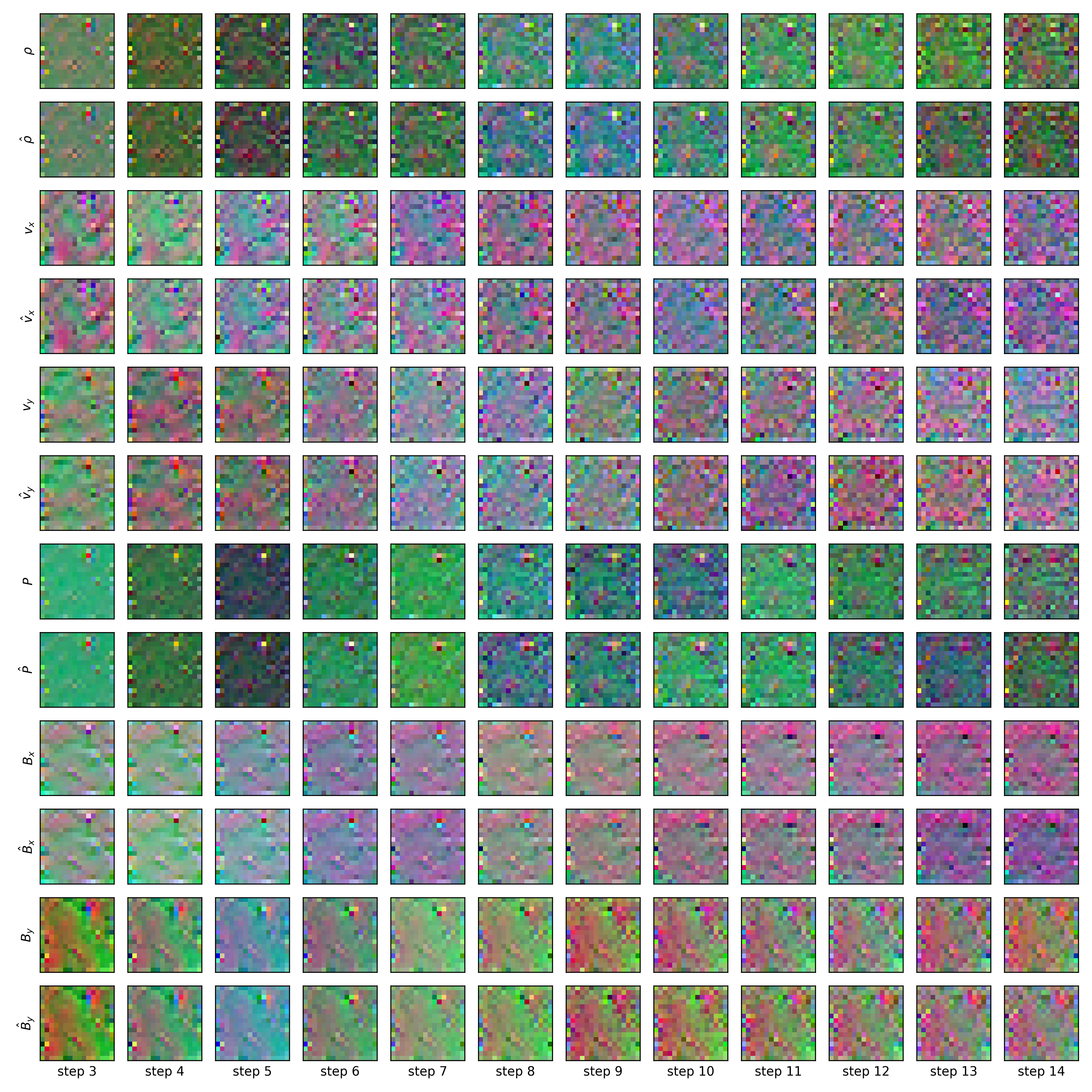}
\caption{Forward roll-out latent space prediction examples.}
\label{fig:forward_roll}
\end{figure*}

\begin{figure*}[ht]
\centering
\includegraphics[width=1.0\linewidth]{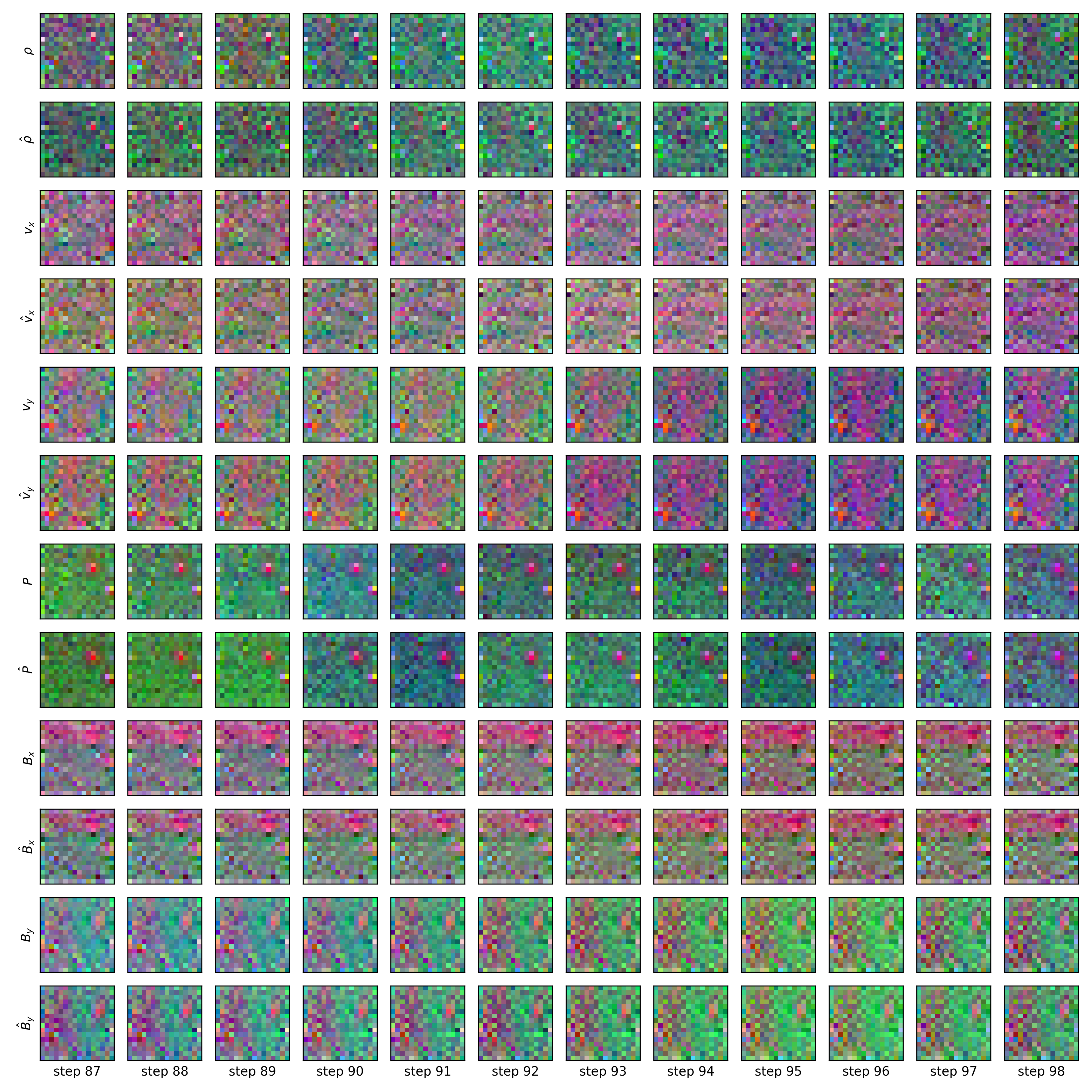}
\caption{Backward roll-out latent space prediction examples.}
\label{fig:backward_roll}
\end{figure*}

We then test the roll-out ability of our overall setup both forward and backward in time. Figures \ref{fig:forward_roll} and \ref{fig:backward_roll} show the first 15 time-steps of both forward and backward latent space roll outs for all 6 MHD fields compared to their true values, respectively.

\begin{figure}[ht]
\centering
\includegraphics[width=1.0\linewidth]{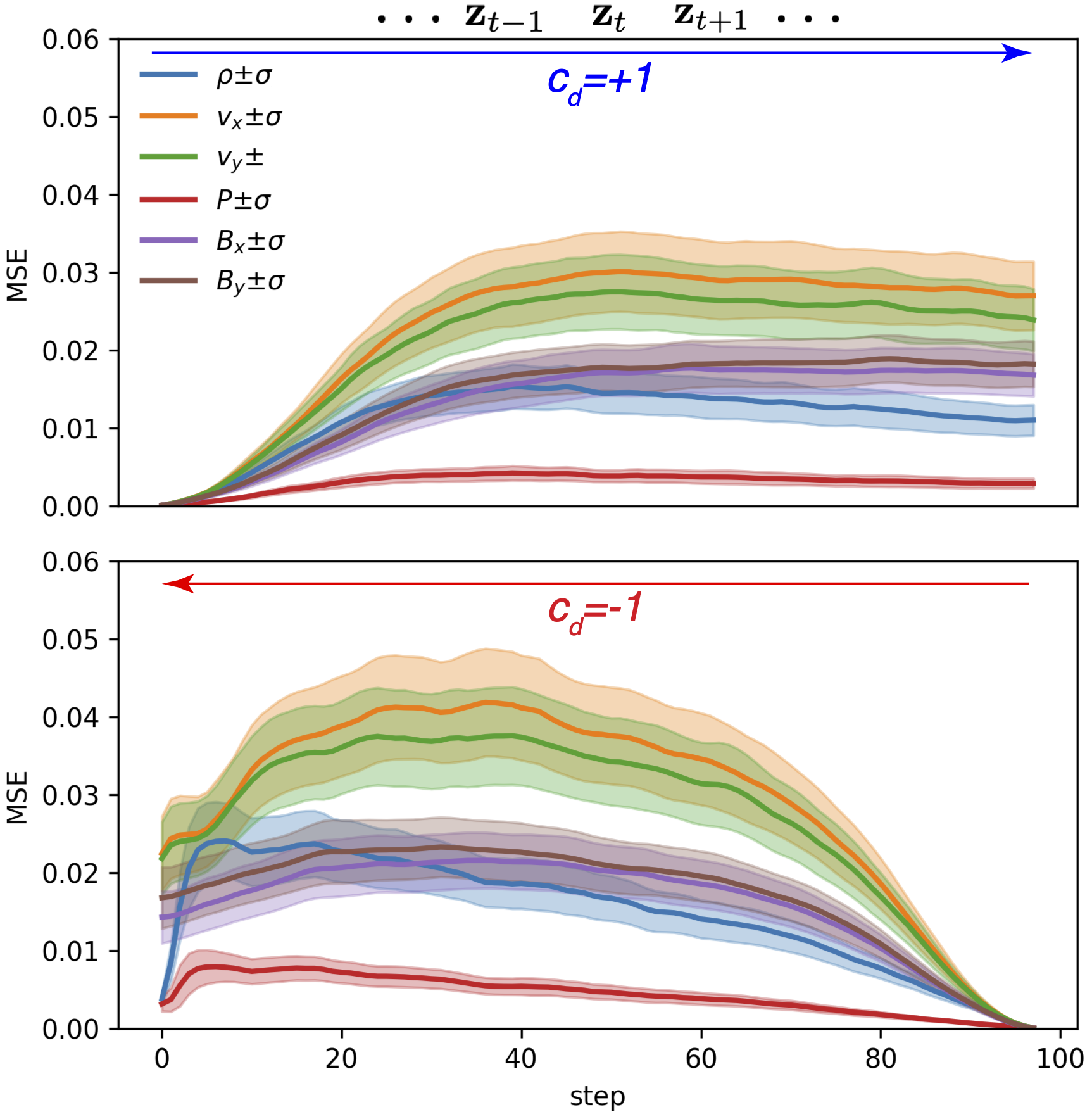}
\caption{Mean error values and their standard deviations are shown for both forward and backward roll outs, averaged over the entire test data set.}
\label{fig:roll_error}
\end{figure}

In either direction, we expect to see that error is accumulated as we autoregressively produce the MHD fields with only the first two time-steps using ground truth latent field representations as inputs and subsequent predictions having to use the diffusion model's imperfect predictions as their own inputs. This is quantified in Figure \ref{fig:roll_error} where we see the mean error values and their standard deviations, for both forward and backward roll outs, averaged over the entire test data set.

\begin{figure*}[ht]
\centering
\includegraphics[width=1.0\linewidth]{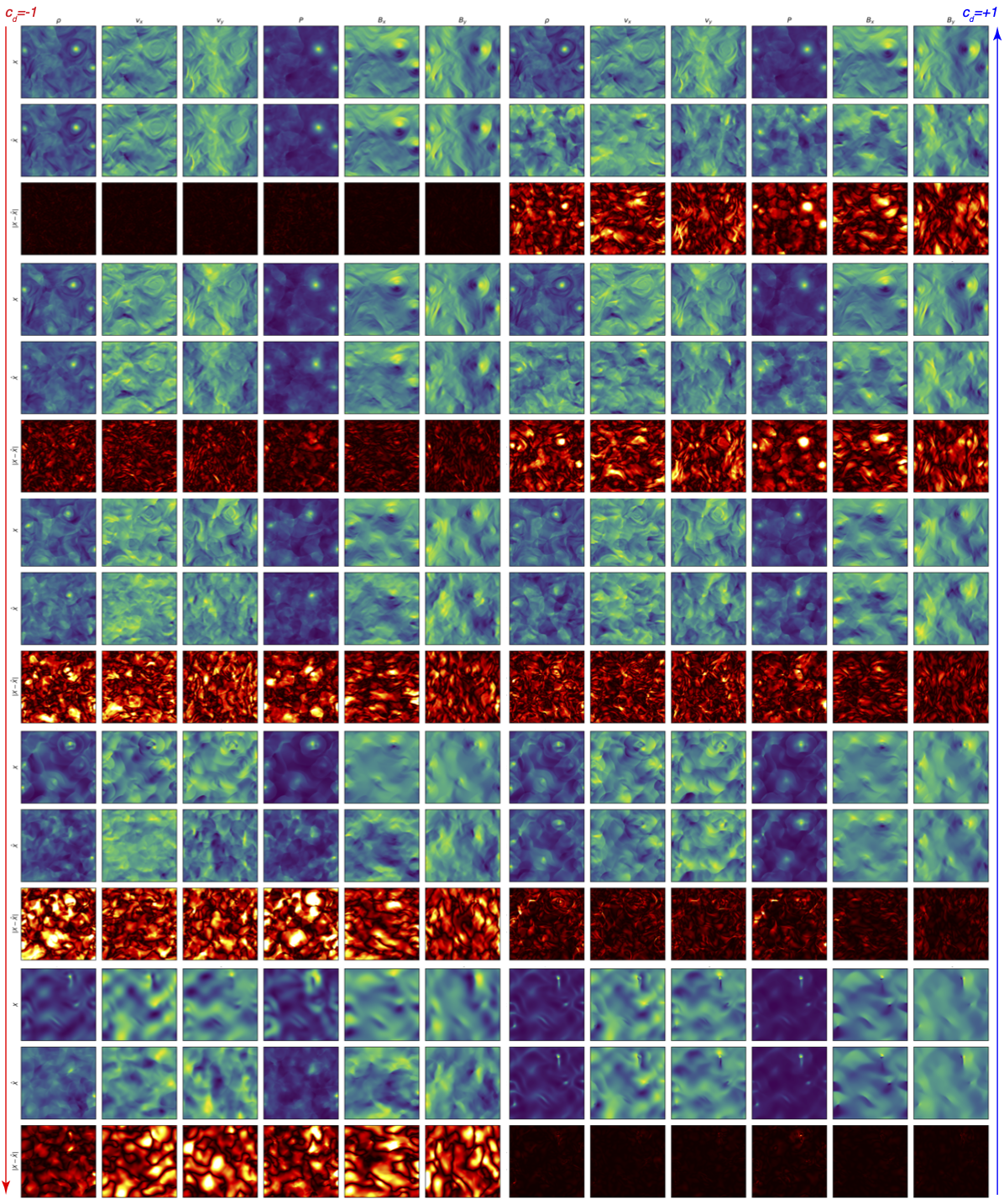}
\caption{Example of predicted reconstructions and true fields shown for autoregressively predicted roll outs of length 2, 20, 40, 50, and 98 time-steps, going backward in time (left) and forward in time (right) for the same test data set.}
\label{fig:roll_error_recon}
\end{figure*}

Figure \ref{fig:roll_error_recon} shows predicted reconstructed and true fields for autoregressively predicted roll outs of length 2, 20, 40, 50, and 98 time-steps, going backward in time (left) and forward in time (right) for a test data set.

%%%%%%%%%%%%%%%%%%%%%%%%%%%%%%%%%%%%%%%%%%%%%%%%%%%%%%%%%%%%%%%%%%
%%%%%%%%%%%%%%%%%%%%%%%%%%%%%%%%%%%%%%%%%%%%%%%%%%%%%%%%%%%%%%%%%%
%%%%%%%%%%%%%%%%%%%%%%%%%%%%%%%%%%%%%%%%%%%%%%%%%%%%%%%%%%%%%%%%%%
%%%%%%%%%%%%%%%%%%%%%%%%%%%%%%%%%%%%%%%%%%%%%%%%%%%%%%%%%%%%%%%%%%
\section{Conclusions}
%%%%%%%%%%%%%%%%%%%%%%%%%%%%%%%%%%%%%%%%%%%%%%%%%%%%%%%%%%%%%%%%%%
%%%%%%%%%%%%%%%%%%%%%%%%%%%%%%%%%%%%%%%%%%%%%%%%%%%%%%%%%%%%%%%%%%
%%%%%%%%%%%%%%%%%%%%%%%%%%%%%%%%%%%%%%%%%%%%%%%%%%%%%%%%%%%%%%%%%%
%%%%%%%%%%%%%%%%%%%%%%%%%%%%%%%%%%%%%%%%%%%%%%%%%%%%%%%%%%%%%%%%%%
In future work, we plan on extending these results to include hard physics constraints on various aspects of the MHD generative process. One approach is the use of physics constrained neural networks (PCNNs), which utilize a novel neural operator approach for building hard physics constraints into convolutional neural networks by simulating potentials (such as the vector potential $\mathbf{A}$) from which electromagnetic fields are then defined according to Maxwell's equations ($\mathbf{B}=\nabla\times\mathbf{A}$), thus guaranteeing hard constraints such as $\nabla\cdot\mathbf{B}=0$, as developed in \cite{scheinker2023physics}.

%%%%%%%%%%%%%%%%%%%%%%%%%%%%%%%%%%%%%%%%%%%%%%%%%%%%%%%%%%%%%%%%%%
%%%%%%%%%%%%%%%%%%%%%%%%%%%%%%%%%%%%%%%%%%%%%%%%%%%%%%%%%%%%%%%%%%
\begin{acknowledgments}
This work was carried out using Google's TPUs, supported by the Google TPU Builders Program.
\end{acknowledgments}
%%%%%%%%%%%%%%%%%%%%%%%%%%%%%%%%%%%%%%%%%%%%%%%%%%%%%%%%%%%%%%%%%%
%%%%%%%%%%%%%%%%%%%%%%%%%%%%%%%%%%%%%%%%%%%%%%%%%%%%%%%%%%%%%%%%%%

\nocite{}
%%%%%%%%%%%%%%%%%%%%%%%%%%%%%%%%%%%%%%%%%%%%%%%%%%%%%%%%%%%%%%%%%%
\bibliography{aipsamp}% Produces the bibliography via BibTeX.

@PREAMBLE{
 "\providecommand{\noopsort}[1]{}" 
 # "\providecommand{\singleletter}[1]{#1}%" 
}

@article{papamakarios2021normalizing,
  title={Normalizing flows for probabilistic modeling and inference},
  author={Papamakarios, George and Nalisnick, Eric and Rezende, Danilo Jimenez and Mohamed, Shakir and Lakshminarayanan, Balaji},
  journal={Journal of Machine Learning Research},
  volume={22},
  number={57},
  pages={1--64},
  year={2021}
}

@article{lipman2022flow,
  title={Flow matching for generative modeling},
  author={Lipman, Yaron and Chen, Ricky TQ and Ben-Hamu, Heli and Nickel, Maximilian and Le, Matt},
  journal={arXiv preprint arXiv:2210.02747},
  year={2022}
}

@misc{song2023consistency,
      title={Consistency Models}, 
      author={Yang Song and Prafulla Dhariwal and Mark Chen and Ilya Sutskever},
      year={2023},
      eprint={2303.01469},
      archivePrefix={arXiv},
      primaryClass={cs.LG},
      url={https://arxiv.org/abs/2303.01469}, 
}

@article{salazar2024posterior,
  title={Posterior Mean Matching: Generative Modeling through Online Bayesian Inference},
  author={Salazar, Sebastian and Kucer, Michal and Wang, Yixin and Casleton, Emily and Blei, David},
  journal={arXiv preprint arXiv:2412.13286},
  year={2024}
}

@article{lecun2006tutorial,
  title={A tutorial on energy-based learning},
  author={LeCun, Yann and Chopra, Sumit and Hadsell, Raia and Ranzato, M and Huang, Fujie and others},
  journal={Predicting structured data},
  volume={1},
  number={0},
  year={2006}
}

@article{peyre2019computational,
  title={Computational optimal transport: With applications to data science},
  author={Peyr{\'e}, Gabriel and Cuturi, Marco and others},
  journal={Foundations and Trends{\textregistered} in Machine Learning},
  volume={11},
  number={5-6},
  pages={355--607},
  year={2019},
  publisher={Now Publishers, Inc.}
}

@article{anderson1982reverse,
  title={Reverse-time diffusion equation models},
  author={Anderson, Brian DO},
  journal={Stochastic Processes and their Applications},
  volume={12},
  number={3},
  pages={313--326},
  year={1982},
  publisher={Elsevier}
}

@inproceedings{scheinker2026phaseflow4d,
  title={Phase{F}low4{D}: {P}hysically Constrained 4{D} Beam Reconstruction via Feedback-Guided Latent Diffusion},
  author={Scheinker, Alexander and Plastun, Alexander and Ostroumov, Peter},
  booktitle={Proceedings of the IEEE/CVF Conference on Computer Vision and Pattern Recognition},
  pages={4729--4737},
  year={2026},
  url={https://openaccess.thecvf.com/content/CVPR2026W/GenRecon3d/html/Scheinker_PhaseFlow4D_Physically_Constrained_4D_Beam_Reconstruction_via_Feedback-Guided_Latent_Diffusion_CVPRW_2026_paper.html}
}

@article{kube2022near,
  title={Near real-time streaming analysis of big fusion data},
  author={Kube, Ralph and Churchill, R Michael and Chang, CS and Choi, Jong and Wang, R and Klasky, Scott and Stephey, Laurie and Dart, E and Choi, MJ},
  journal={Plasma Physics and Controlled Fusion},
  volume={64},
  number={3},
  pages={035015},
  year={2022},
  publisher={IOP Publishing}
}

@inproceedings{choi2020data,
  title={Data federation challenges in remote near-real-time fusion experiment data processing},
  author={Choi, Jong and Wang, Ruonan and Churchill, R Michael and Kube, Ralph and Choi, Minjun and Park, Jinseop and Logan, Jeremy and Mehta, Kshitij and Eisenhauer, Greg and Podhorszki, Norbert and others},
  booktitle={Smoky Mountains Computational Sciences and Engineering Conference},
  pages={285--299},
  year={2020},
  organization={Springer}
}

@article{scheinker2023physics,
  title={Physics-constrained 3{D} convolutional neural networks for electrodynamics},
  author={Scheinker, Alexander and Pokharel, Reeju},
  journal={APL Machine Learning},
  volume={1},
  number={2},
  year={2023},
  publisher={AIP Publishing},
  url={https://doi.org/10.1063/5.0132433}
}

@article{orszag1979small,
  title={Small-scale structure of two-dimensional magnetohydrodynamic turbulence},
  author={Orszag, Steven A and Tang, Cha-Mei},
  journal={Journal of Fluid Mechanics},
  volume={90},
  number={1},
  pages={129--143},
  year={1979},
  publisher={Cambridge University Press}
}

@article{baranov1991mhd,
  title={MHD method of measuring high-power ion beam parameters},
  author={Baranov, SV},
  journal={Sov. J. Plasma Phys},
  volume={17},
  pages={156--157},
  year={1991}
}

@inproceedings{gill2007development,
  title={Development of a diagnostic array for the measurement of velocity profiles across open-channel liquid metal flows},
  author={Gill, Alexander and Nornberg, Mark and Ji, Hantao and Peterson, Jayson Luc},
  booktitle={APS Division of Plasma Physics Meeting Abstracts},
  volume={49},
  pages={JP8--021},
  year={2007}
}

@article{smolentsev2021physical,
  title={Physical background, computations and practical issues of the magnetohydrodynamic pressure drop in a fusion liquid metal blanket},
  author={Smolentsev, Sergey},
  journal={Fluids},
  volume={6},
  number={3},
  pages={110},
  year={2021},
  publisher={MDPI}
}

@article{arefiev2005magnetohydrodynamic,
  title={Magnetohydrodynamic scenario of plasma detachment in a magnetic nozzle},
  author={Arefiev, Alexey V and Breizman, Boris N},
  journal={physics of plasmas},
  volume={12},
  number={4},
  year={2005},
  publisher={AIP Publishing}
}

@article{heidbrink2008basic,
  title={Basic physics of Alfv{\'e}n instabilities driven by energetic particles in toroidally confined plasmas},
  author={Heidbrink, WW},
  journal={Physics of Plasmas},
  volume={15},
  number={5},
  year={2008},
  publisher={AIP Publishing}
}

@article{izzo2008magnetohydrodynamic,
  title={Magnetohydrodynamic simulations of massive gas injection into Alcator C-Mod and DIII-D plasmas},
  author={Izzo, VA and Whyte, DG and Granetz, RS and Parks, PB and Hollmann, EM and Lao, LL and Wesley, JC},
  journal={Physics of Plasmas},
  volume={15},
  number={5},
  year={2008},
  publisher={AIP Publishing}
}

@article{jardin2022ideal,
  title={Ideal MHD limited electron temperature in spherical tokamaks},
  author={Jardin, Stephen C and Ferraro, Nathaniel M and Guttenfelder, Walter and Kaye, Stanley M and Munaretto, Stefano},
  journal={Physical review letters},
  volume={128},
  number={24},
  pages={245001},
  year={2022},
  publisher={APS}
}

@article{orain2013non,
  title={Non-linear magnetohydrodynamic modeling of plasma response to resonant magnetic perturbations},
  author={Orain, F and B{\'e}coulet, M and Dif-Pradalier, G and Huijsmans, G and Pamela, S and Nardon, E and Passeron, C and Latu, G and Grandgirard, V and Fil, A and others},
  journal={Physics of Plasmas},
  volume={20},
  number={10},
  year={2013},
  publisher={AIP Publishing}
}

@article{korving2023development,
  title={Development of the neutral model in the nonlinear MHD code JOREK: Application to E$\times$ B drifts in ITER PFPO-1 plasmas},
  author={Korving, SQ and Huijsmans, GTA and Park, J-S and Loarte, A and Jorek Team and others},
  journal={Physics of Plasmas},
  volume={30},
  number={4},
  year={2023},
  publisher={AIP Publishing}
}

@article{haas2005magnetohydrodynamic,
  title={A magnetohydrodynamic model for quantum plasmas},
  author={Haas, F},
  journal={Physics of Plasmas},
  volume={12},
  number={6},
  year={2005},
  publisher={AIP Publishing}
}

@article{cho2002compressible,
  title={Compressible sub-Alfv{\'e}nic MHD turbulence in low-$\beta$ plasmas},
  author={Cho, Jungyeon and Lazarian, Alex},
  journal={Physical Review Letters},
  volume={88},
  number={24},
  pages={245001},
  year={2002},
  publisher={APS}
}

@article{matthaeus2021turbulence,
  title={Turbulence in space plasmas: Who needs it?},
  author={Matthaeus, WH},
  journal={Physics of Plasmas},
  volume={28},
  number={3},
  year={2021},
  publisher={AIP Publishing}
}

@article{alfven1942existence,
  title={Existence of electromagnetic-hydrodynamic waves},
  author={Alfv{\'e}n, Hannes},
  journal={nature},
  volume={150},
  number={3805},
  pages={405--406},
  year={1942},
  publisher={Nature Publishing Group UK London}
}

@article{scheinker2014hardware,
  title={In-hardware demonstration of model-independent adaptive tuning of noisy systems with arbitrary phase drift},
  author={Scheinker, Alexander and Baily, Scott and Young, Daniel and Kolski, Jeffrey S and Prokop, Mark},
  journal={Nuclear Instruments and Methods in Physics Research Section A: Accelerators, Spectrometers, Detectors and Associated Equipment},
  volume={756},
  pages={30--38},
  year={2014},
  publisher={Elsevier}
}

@article{scheinker2013extremum,
  title={Extremum seeking-based optimization of high voltage converter modulator rise-time},
  author={Scheinker, Alexander and Bland, Michael and Krsti{\'c}, Miroslav and Audia, Jeff},
  journal={IEEE Transactions on Control Systems Technology},
  volume={22},
  number={1},
  pages={34--43},
  year={2013},
  publisher={IEEE}
}

@article{mocz2014constrained,
  title={A constrained transport scheme for MHD on unstructured static and moving meshes},
  author={Mocz, Philip and Vogelsberger, Mark and Hernquist, Lars},
  journal={Monthly Notices of the Royal Astronomical Society},
  volume={442},
  number={1},
  pages={43--55},
  year={2014},
  publisher={Oxford University Press}
}

@article{mocz2016moving,
  title={A moving mesh unstaggered constrained transport scheme for magnetohydrodynamics},
  author={Mocz, Philip and Pakmor, R{\"u}diger and Springel, Volker and Vogelsberger, Mark and Marinacci, Federico and Hernquist, Lars},
  journal={Monthly Notices of the Royal Astronomical Society},
  volume={463},
  number={1},
  pages={477--488},
  year={2016},
  publisher={The Royal Astronomical Society}
}

@misc{mocz2023ct_code,
  author       = {Mocz, Philip},
  title        = {{constrainedtransport-python}: Finite Volume Constrained
                  Transport Simulation of the {Orszag}--{Tang} Vortex},
  year         = {2023},
  howpublished = {\url{https://github.com/pmocz/constrainedtransport-python}},
  note         = {GitHub repository. Accessed: 2026-06-26}
}

@misc{mocz2023ct_article,
  author       = {Mocz, Philip},
  title        = {Create Your Own Constrained Transport Magnetohydrodynamics
                  Simulation (With {Python})},
  year         = {2023},
  howpublished = {\url{https://levelup.gitconnected.com/create-your-own-constrained-transport-magnetohydrodynamics-simulation-with-python-276f787f537d}},
  note         = {Level Up Coding, Medium. Accessed: 2026-06-26}
}

@inproceedings{scheinker2013model,
  title={Model independent beam tuning},
  author={Scheinker, Alexander and others},
  booktitle={Proceedings of the 2013 International Particle Accelerator Conference, Shanghai, China},
  year={2013},
  url={https://proceedings.jacow.org/IPAC2013/papers/tupwa068.pdf}
}

@article{scheinker2016bounded,
  title={Bounded extremum seeking with discontinuous dithers},
  author={Scheinker, Alexander and Scheinker, David},
  journal={Automatica},
  volume={69},
  pages={250--257},
  year={2016},
  publisher={Elsevier},
  url={https://doi.org/10.1016/j.automatica.2016.02.023}
}

@article{scheinker2024cdvae,
  title={c{DVAE}: {VAE}-guided diffusion for particle accelerator beam 6D phase space projection diagnostics},
  author={Scheinker, Alexander},
  journal={Scientific Reports},
  volume={14},
  number={1},
  pages={29303},
  year={2024},
  publisher={Nature Publishing Group UK London},
  url={https://www.nature.com/articles/s41598-024-80751-1}
}

@inproceedings{de2022event,
  title={Event-driven adaptive Vertical Stabilization in tokamaks based on a bounded Extremum Seeking algorithm},
  author={De Tommasi, Gianmaria and Dubbioso, Sara and Mele, Adriano and Pironti, Alfredo},
  booktitle={2022 IEEE Conference on Control Technology and Applications (CCTA)},
  pages={831--836},
  year={2022},
  organization={IEEE}
}

@article{saxena2025improved,
  title={Improved Robustness of Deep Reinforcement Learning for Control of Time-Varying Systems by Bounded Extremum Seeking},
  author={Saxena, Shaifalee and Williams, Alan and Fierro, Rafael and Scheinker, Alexander},
  journal={arXiv preprint arXiv:2510.02490},
  year={2025},
  url={https://arxiv.org/abs/2510.02490}
}

@article{saxena2026deep,
  title={Deep Reinforcement Learning for Robotic Manipulation under Distribution Shift with Bounded Extremum Seeking},
  author={Saxena, Shaifalee and Fierro, Rafael and Scheinker, Alexander},
  journal={arXiv preprint arXiv:2604.01142},
  year={2026},
  url={https://arxiv.org/abs/2604.01142}
}

@article{saxena2026mahalanobis,
  title={Mahalanobis-{G}uided Latent {OOD} Detection for Hybrid {ES}-{DRL} Control in Time-Varying Systems},
  author={Saxena, Shaifalee and Scheinker, Alexander},
  journal={arXiv preprint arXiv:2606.11474},
  year={2026},
  url={https://arxiv.org/abs/2606.11474}
}
%%%%%%%%%%%%%%%%%%%%%%%%%%%%%%%%%%%%%%%%%%%%%%%%%%%%%%%%%%%%%%%%%%
\end{document}